\documentclass[10pt,journal,compsoc]{IEEEtran}

\ifCLASSOPTIONcompsoc
  \usepackage[nocompress]{cite}
\else
  \usepackage{cite}
\fi

\usepackage{amsmath,amssymb,amsfonts}

\usepackage{algorithm}        
\usepackage[noend]{algpseudocode} 
\usepackage{graphicx}
\usepackage{textcomp}
\usepackage{xcolor}
\usepackage{placeins}
\usepackage{float}
\usepackage{tabularx}
\usepackage{adjustbox}
\usepackage{caption} 
 \usepackage{multirow}
 \usepackage{lscape}
\usepackage{algorithm}
\usepackage{algpseudocode}
\usepackage{graphicx}
\usepackage{tabularx,booktabs}
\usepackage{lscape}
\usepackage{longtable}
\usepackage{makecell}   

\newcolumntype{C}{>{\centering\arraybackslash}X} 
\usepackage{amsmath}
\usepackage{amssymb}

\usepackage{pifont}
\newcommand{\cmark}{\ding{51}}%
\newcommand{\xmark}{\ding{55}}%
\newcolumntype{Y}{>{\raggedright\arraybackslash}X}
\newcolumntype{C}{>{\centering\arraybackslash}p{0.9cm}} 
\newcolumntype{C}[1]{>{\centering\arraybackslash}m{#1}}

\def\BibTeX{{\rm B\kern-.05em{\sc i\kern-.025em b}\kern-.08em
    T\kern-.1667em\lower.7ex\hbox{E}\kern-.125emX}}

\ifCLASSINFOpdf
\else
\fi

\begin{document}
%
\title{Large Language Models Hallucination: A Comprehensive Survey}
%
%
%
%

\author{Aisha Alansari,~\IEEEmembership{KFUPM}
        and~Hamzah Luqman,~\IEEEmembership{KFUPM} \\ aisha.ansari@kfupm.edu.sa, hluqman@kfupm.edu.sa}

\IEEEtitleabstractindextext{%
\begin{abstract} 
Large language models (LLMs) have transformed natural language processing, achieving remarkable performance across diverse tasks. However, their impressive fluency often comes at the cost of producing false or fabricated information, a phenomenon known as hallucination. Hallucination refers to the generation of content by an LLM that is fluent and syntactically correct but factually inaccurate or unsupported by external evidence.  Hallucinations undermine the reliability and trustworthiness of LLMs, especially in domains requiring factual accuracy. This survey provides a comprehensive review of research on hallucination in LLMs, with a focus on causes, detection, and mitigation. We first present a taxonomy of hallucination types and analyze their root causes across the entire LLM development lifecycle, from data collection and architecture design to inference. We further examine how hallucinations emerge in key natural language generation tasks. Building on this foundation, we introduce a structured taxonomy of detection approaches and another taxonomy of mitigation strategies. We also analyze the strengths and limitations of current detection and mitigation approaches and review existing evaluation benchmarks and metrics used to quantify LLMs hallucinations. Finally, we outline key open challenges and promising directions for future research, providing a foundation for the development of more truthful and trustworthy LLMs.
\end{abstract}

\begin{IEEEkeywords}
LLMs, Hallucination, Hallucination Causes, Hallucination Detection, Hallucination Mitigation, Hallucination Benchmarks, Hallucination Metrics
\end{IEEEkeywords}}

\maketitle

\section{Introduction}
\label{sec_intro}
Natural language generation (NLG) has significantly advanced in recent years due to the progress in transformer-based language models (LMs). Large language models (LLMs), such as ChatGPT \cite{openai2023gpt}, Claude \cite{claude}, and Bard \cite{bard}, have revolutionized natural language processing by enabling powerful capabilities across a diverse range of applications. These models have offered substantial enhancements in efficiency and productivity, which have enhanced progress in downstream tasks, such as question answering (QA), abstractive summarization, dialogue generation, and data-to-text generation. Despite these breakthroughs, a critical challenge has emerged with LLMs, known as hallucination.


Hallucination refers to the generation of content by LLMs that is fluent and syntactically correct, but factually inaccurate or unsupported by external evidence \cite{xu2024hallucination,ji2023survey}. It can result in significant repercussions, including the dissemination of misinformation and breaches of privacy. In contrast to conventional artificial intelligence (AI) systems, which are usually trained in data related to a specific task, LLMs have been trained using large amounts of online textual data \cite{brown2020language}. This broad coverage enables exceptional coherence and fluency; however, it also increases the risk of inaccuracies.
LLMs can reflect biases present in their training data, misinterpret unclear prompts, or modify information to align with the perceived intent of input \cite{bender2021dangers}. This is particularly concerning when individuals depend on language generation capabilities for sensitive applications, such as summarizing medical data, customer service dialogues, financial analysis reports, and providing legal counsel.


Investigating the stages of LLM development helps to understand the root causes of hallucinations throughout their pre-training to the generation pathway. It also provides guidance in developing hallucination detection and mitigation techniques for LLMs. Based on the standard stages of LLMs' development, we examine each stage of the development pipeline to identify the factors that contribute to hallucinations. We divide the LLM development pipeline into six distinct stages: data collection and preparation, model architecture, pre-training, fine-tuning, evaluation, and inference. This examination facilitates a thorough understanding of the underlying causes of hallucinations at every stage. 

Moreover, we propose a taxonomy for hallucination detection techniques. This taxonomy categorizes hallucination detection techniques into retrieval-, uncertainty-, embedding-, learning-, and self-consistency-based techniques. Based on our findings, it is challenging for a single hallucination detection approach to perform well under all circumstances. Retrieval-based detection approaches effectively deal with factual hallucinations but are extremely sensitive to the quality of external knowledge. Similarly, learning-based detection approaches are accurate but depend on high-quality annotated data. Uncertainty-based detection addresses the challenge of data dependency using model confidence rather than external labeled datasets. However, its effectiveness is highly sensitive to the calibration of uncertainty thresholds, and it often fails to detect hallucinations when the model shows high confidence in an incorrect response. Similarly, self-consistency detection approaches can detect logical and contextual inconsistencies without relying on external evidence. However, these approaches struggle with subtle factual errors and are highly dependent on prompt diversity and sampling strategies. Embedding-based detection techniques are robust in capturing semantic discrepancies. Nevertheless, their performance in detecting hallucination can be degraded in out-of-domain data and low-resource languages. Therefore, the combination of complementary approaches (e.g., learning with uncertainty or retrieval with learning) is a promising direction to improve the overall detection robustness and accuracy.
 
Furthermore, we expanded the existing taxonomy of hallucination mitigation techniques derived from prior research \cite{tonmoy2024comprehensive,huang2023survey,sahoo2024comprehensive} by categorizing them into four categories: prompt-, retrieval-, reasoning-, and model-centric training and adaptation-based approaches. Prompt-based mitigation approaches depend on structured prompting strategies to guide models towards generating factual content. Retrieval-based mitigation methods depend on external knowledge to ground outputs. Reasoning-based mitigation techniques, such as chain-of-thought prompting (CoT) and self-consistency, enhance logical coherence and internal consistency. Model-centric training and adaptation-based approaches involve modifying architectures, adjusting training objectives, and employing fine-tuning procedures to improve models' intrinsic factuality and reliability. Based on our analysis, we show that no single approach completely mitigates hallucination. Consequently, more effective mitigation is a combination of complementary techniques. The most promising are hybrid approaches that combine prompting or reasoning-based techniques with retrieval-based and model-centric training and adaptation strategies.

Moreover, we discuss the challenges faced by current hallucination detection and mitigation methodologies, and propose potential future work directions for detecting and mitigating hallucinations in LLMs. While prior surveys laid critical groundwork, this survey builds upon previous research by providing a comprehensive analysis of the causes of hallucinations and techniques that have been proposed for hallucination detection and mitigation. The main contributions of this survey can be summarized as follows:
\begin{itemize}
    \item \textbf{Hallucination causes analysis:} This survey presents a deep analysis of hallucination causes across all stages of the LLM development cycle, from data collection and architecture design to inference.
    \item \textbf{LLMs hallucination taxonomy:} A comprehensive taxonomy is proposed in this survey for the hallucination causes, and state-of-the-art (SOTA) approaches that have been proposed for hallucination detection and mitigation.
    \item \textbf{Hallucination detection discussion:} We propose a structured classification of hallucination detection methods by categorizing them into five main categories: retrieval, uncertainty, embedding, learning, and self-consistency-based detection approaches. Each category have been discussed deeply to show its potential for hallucination detection.
    \item \textbf{Hallucination mitigation discussion:} The hallucination mitigation methods is classified in this survey into four main categories: prompt, retrieval, reasoning, and model-centric training and adaptation-based techniques. Each category have been discussed deeply to show its potential for hallucination mitigation.
    \textcolor{black}{\item \textbf{Hallucination explainability discussion:} The hallucination explainability methods is classified in this survey into model-derived and grounding explainability. Each category includes recent studies to show their potential for hallucination explainability.}
    \item \textbf{Datasets and Evaluation Metrics:} We review the benchmark datasets used for hallucination detection and mitigation and identify their limitations. We also discussed the metrics used to evaluate detection and mitigation techniques.  
    \item \textbf{In-depth Analysis of Reasoning-Aware Solutions:} We provide an in-depth review of recent reasoning-based mitigation methods, including CoT, iterative refinement, and chain-of-verification techniques, highlighting their roles in reducing hallucination in complex tasks.
    \item \textbf{Multilingual and Low-Resource Emphasis:} This survey identifies challenges unique to underrepresented languages and surveys cross-lingual transfer, multilingual fine-tuning, and prompt adaptation techniques to mitigate hallucinations in low-resource settings.
    
\end{itemize}

The remainder of the survey is organized as follows: Section \ref{sec_related_surveys} provides a review of related surveys. Then, a detailed discussion about hallucination, its types, and how it appears in diverse NLG tasks is presented in Section \ref{sec_LLMHallu}.  
 Section \ref{sec:sources} presents a comprehensive taxonomy of the reasons for hallucination across all stages of the LLM development cycle, followed by a detailed classification of detection techniques (Section \ref{sec_detection}) and mitigation strategies (Section \ref{sec_mitigation}). The datasets used to train and evaluate hallucination detection and mitigation techniques are presented in Section \ref{sec_benchmark}, and the evaluation metrics are discussed in Section \ref{sec_metric}. Finally, Section \ref{sec_open_issues} presents a detailed discussion of the open issues and future research directions, and Section \ref{sec:conclusion} concludes this study.

\section{Related Surveys} 
\label{sec_related_surveys}
Given the rapid evolution of LLMs and their expanding applications across diverse domains, hallucination detection and mitigation techniques have also progressed. Consequently, there is a need to thoroughly analyze the causes of hallucinations, their detection, and their mitigation techniques to remain up to date. Investigating these factors will contribute to diagnosing the root causes of LLM hallucinations and developing more effective detection and mitigation techniques. Several surveys have been published recently on LLMs hallucination \cite{ji2023survey,huang2023survey,tonmoy2024comprehensive,zhang2023siren,ye2023cognitive}. These surveys examine LLMs' hallucination from multiple points of view and provide significant insights. Ji et al. \cite{ji2023survey} reviewed hallucination across various NLG tasks and outlined the mitigation methods applied and the evaluation metrics used in each task. Ye et al. \cite{ye2023cognitive} extended this work by proposing a new taxonomy for the detection and mitigation of hallucination. Zhang et al. \cite{zhang2023siren} highlighted some issues related to LLMs, such as input, context, and fact-conflicting hallucinations.
Tonmoy et al. \cite{tonmoy2024comprehensive} focused on hallucination mitigation by presenting a taxonomy that categorizes mitigation techniques into prompt engineering, retrieval-augmented generation (RAG), self-refinement, and decoding strategies. Huang et al. \cite{huang2023survey} presented a dual taxonomy of factuality and faithfulness hallucinations and defined hallucination causes in the data, training, and inference stages. The survey also linked detection and mitigation strategies to their foundational causes to guide robust system development.  

More recently, Saxena and Bhattacharyya \cite{saxena2024hallucination} provided a comprehensive survey on hallucination detection methods, classifying hallucinations into intrinsic and extrinsic types. In another study, Cossio \cite{cossio2025comprehensive} offered a taxonomy of hallucination types, including factual errors, contextual inconsistencies, temporal disorientation, ethical violations, and domain-specific hallucinations, while categorizing causes into data, model, and prompt-related factors. Malin et al. \cite{malin2025review} reviewed faithfulness metrics used to evaluate hallucinations across summarization, QA, and machine translation. They also linked mitigation strategies, such as RAG and prompting frameworks, with improved faithfulness. Qi et al. \cite{qi2024survey} focused on automatic hallucination evaluation. Rahman et al. \cite{rahman2025hallucination} reviewed fact-checking and factuality evaluation in LLMs, and analyzed how hallucination affects LLM reliability. 

Unlike prior surveys, our survey makes several distinctive contributions as shown in Table \ref{tab:related-surveys}. Compared to previous work, our survey proposes a detailed taxonomy of LLM hallucination causes explicitly linked to the LLM development cycle and develops more fine-grained taxonomies for both hallucination detection and mitigation. We further review these techniques in multilingual contexts, addressing language-specific challenges that are often disregarded in the literature. Our analysis also spans the entire LLM lifecycle, from pre-training and fine-tuning to inference, which provides a more comprehensive review than previous studies. In addition, we discuss the limitations of each detection and mitigation category and present detailed future research directions to guide advancements in this field. Furthermore, we provide an in-depth review of reasoning-based mitigation methods since these approaches represent a newer wave of mitigation techniques that differ fundamentally from prompt engineering, as they focus on structuring and verifying the model’s internal reasoning rather than only modifying input prompts.

\begin{table*}[t]
\centering
\caption{A comparative analysis of existing LLM hallucination surveys.}
\label{tab:related-surveys}
\begin{tabularx}{\textwidth}{@{} l C{1.5cm}
C{1.5cm} C{1.5cm} C{1.5cm} C{1.5cm} C{1.5cm} C{1.0cm} C{1.5cm} C{1.0cm} @{}}
\toprule
\multirow{2}{*}{\textbf{Reference}} & \multirow{2}{*}{\textbf{Year}} &
\multicolumn{8}{c}{\textbf{Features}}\\
\cmidrule(lr){3-10}
& & \textbf{F1} & \textbf{F2} & \textbf{F3} & \textbf{F4} & \textbf{F5} & \textbf{F6} & \textbf{F7} & \textbf{F8} \\
\midrule
\cite{ji2023survey}            & 2023 & \cmark & \xmark & \cmark & \xmark & \cmark & \xmark & \cmark & \xmark \\
\cite{ye2023cognitive}         & 2023 & \xmark & \cmark & \cmark & \cmark & \xmark & \xmark & \cmark & \xmark \\
\cite{zhang2023siren}          & 2023 & \cmark & \xmark & \cmark & \cmark & \cmark & \xmark & \cmark & \xmark \\
\cite{huang2023survey}         & 2023 & \cmark & \cmark & \cmark & \cmark & \xmark & \xmark & \cmark & \xmark \\
\cite{tonmoy2024comprehensive} & 2024 & \xmark & \xmark & \cmark & \xmark & \xmark & \xmark & \cmark & \cmark \\
\cite{saxena2024hallucination} & 2024 & \cmark & \cmark & \xmark & \cmark & \xmark & \xmark & \cmark & \xmark \\
\cite{qi2024survey}            & 2024 & \xmark & \xmark & \xmark & \cmark & \cmark & \cmark & \cmark & \xmark \\
\cite{jiang2024survey}         & 2024 & \xmark & \cmark & \cmark & \xmark & \xmark & \xmark & \cmark & \xmark \\
\cite{rahman2025hallucination} & 2025 & \xmark & \xmark & \cmark & \cmark & \cmark & \xmark & \cmark & \xmark \\
\cite{malin2025review}         & 2025 & \xmark & \xmark & \xmark & \cmark & \cmark & \xmark & \xmark & \xmark \\
\cite{cossio2025comprehensive} & 2025 & \cmark & \xmark & \cmark & \cmark & \cmark & \xmark & \cmark & \xmark \\
\textbf{Our Survey}            & 2025 & \cmark & \cmark & \cmark & \cmark & \cmark & \cmark & \cmark & \cmark \\
\bottomrule
\end{tabularx}
\vspace{2pt}
\raggedright\footnotesize
F1: Analyzes causes of hallucination; 
F2: Proposes a new detection taxonomy; 
F3: Proposes a new mitigation taxonomy; 
F4: Surveys benchmark datasets; 
F5: Surveys evaluation metrics; 
F6: Covers multilingual or cross-lingual hallucination detection and mitigation techniques; 
F7: Provides explicit limitations and future-work directions; 
F8: Surveys reasoning-based techniques.
\end{table*}

\section{LLMs Hallucination}
\label{sec_LLMHallu}

\subsection{Definition of Hallucination}
In a psychological context, the term "hallucination" refers to a false perception of objects or events \cite{fish2009perception}. It involves the experience of seeing, hearing, feeling, smelling, or tasting stimuli that are not actually present. These sensations are often indistinguishable from reality to the one experiencing them \cite{shah20147}. In the context of LLMs, hallucination refers to the generation of text that appears reasonable, fluent, and coherent but lacks grounding in factual or accurate information \cite{maynez2020faithfulness}. 


Hallucinations in earlier LMs, such as n-gram models, were more readily identifiable due to their restricted generative capabilities. Unlike modern LLMs, which produce highly fluent text, earlier LMs often generate nonsensical or clearly inaccurate outputs, which makes their hallucinations easier to detect \cite{koehn2017six}. Conversely, it is more challenging to detect hallucinations in texts generated by modern LLMs, such as GPT \cite{brown2020language} and T5 \cite{raffel2020exploring}, since their capacity to produce fluent and contextually relevant text has markedly improved \cite{maynez2020faithfulness}.

\begin{figure}
    \centering
    \includegraphics[width=1.0\linewidth]{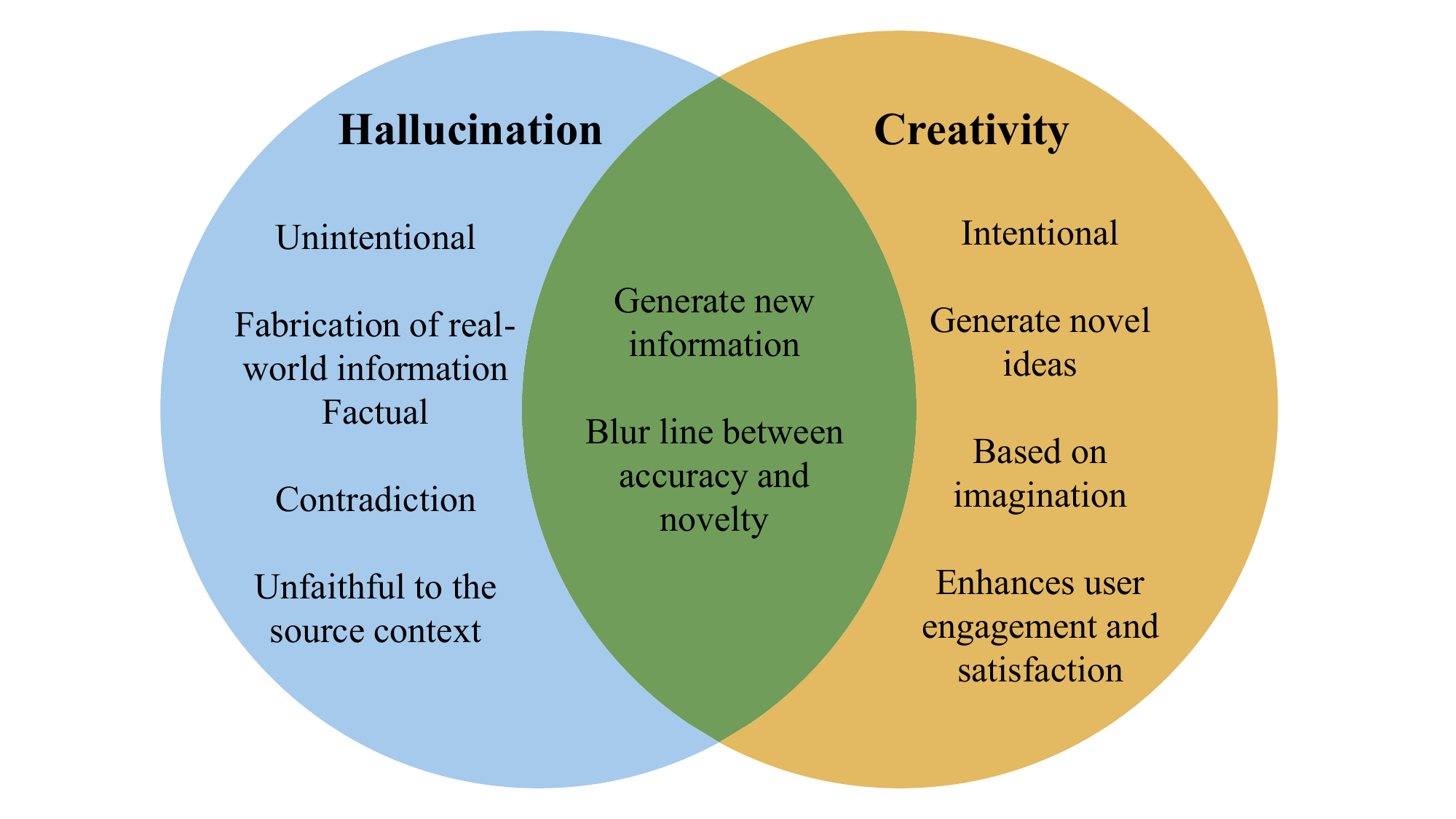}
    \caption{The differences and similarities between hallucination and creativity in LLM outputs.}
    \label{fig:creativity}
\end{figure}

However, it is important to distinguish between creativity and hallucination. Hallucinatory outputs can be valued in creative writing or ideation tasks, which means that there is a gray zone. Creativity allows LLMs to produce novel and imaginative responses, such as poetry or fictional storytelling. In contrast, hallucination involves generating factually incorrect or misleading content. Creativity is known to be deliberate and goal-driven, while hallucinations are typically unintended by the model designer or user \cite{jiang2024survey}. Moreover, creativity does not inherently violate accuracy, whereas hallucination does \cite{jiang2024survey}. Figure \ref{fig:creativity} explains the similarities and differences between creativity and hallucination.

\subsection{Types of Hallucination}
\begin{figure}[h]
    \centering
    \includegraphics[width=1.0\linewidth]{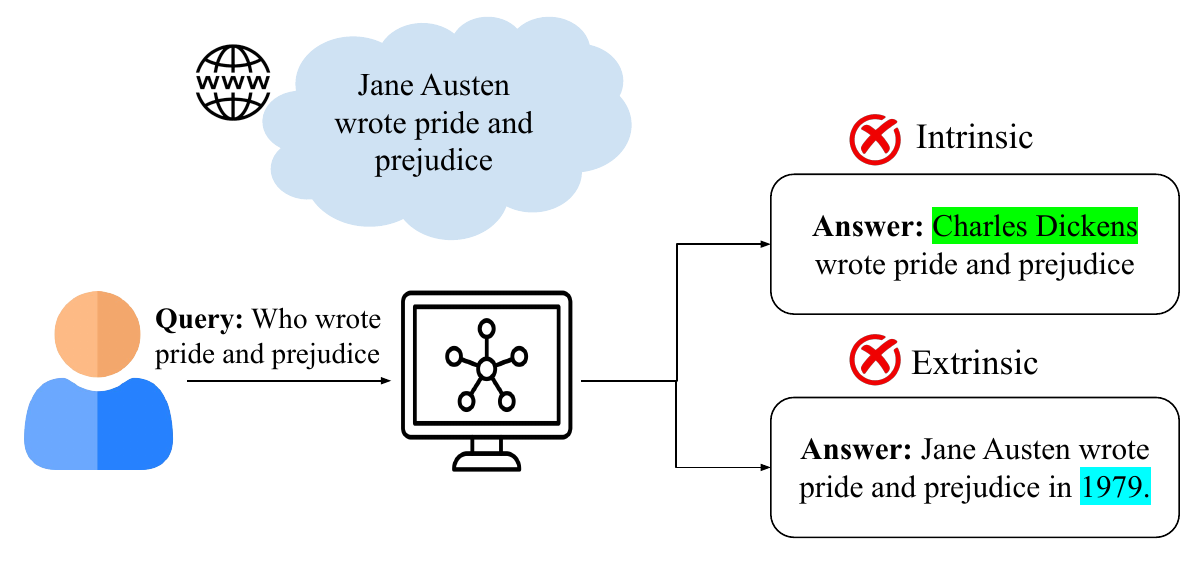}
    \caption{The difference between  \colorbox{green}{intrinsic} and \colorbox{cyan!50}{extrinsic} hallucination.}
    \label{fig:EI}
\end{figure}

Hallucination in NLG can be classified into two types: intrinsic and extrinsic \cite{maynez2020faithfulness}. Intrinsic hallucination occurs when the output of the LLM contradicts facts present in the source document. These errors often arise due to misinterpretation of context, entity confusion, or biases in the training data. Figure \ref{fig:EI} shows an example of intrinsic hallucination. As shown in the figure, the LLM erroneously generates \textit{"Charles Dickens"}, when a user asks about the author of 'Pride and Prejudice', which contradicts the information in the ground truth.  On the other hand, extrinsic hallucinations are characterized by the inclusion of information in the output that is not present in the ground truth. In contrast to intrinsic hallucinations, extrinsic hallucinations introduce additional content that cannot be verified against the source, rather than contradicting it. In such cases, the text appears plausible; however, it lacks explicit support from the input. It is important to note that extrinsic hallucinations can often contain factually accurate information that is derived from external knowledge, despite not being explicitly stated in the source. As shown in Figure \ref{fig:EI}, when a user asks about the author of Pride and Prejudice, the LLM generates the correct answer with an extra entity \textit{"completed the manuscript in 1797"} that is not present in the ground truth data. While this statement may be factually accurate, it is not explicitly provided in the ground truth data. This demonstrates how an LLM might interpolate prior knowledge or generate plausible-sounding additions that lack direct verification.

Huang et al. \cite{huang2023survey} extended the hallucination classification to factual and faithful. Factuality hallucination describes the divergence between produced content and known real-world facts, often appearing as a factual contradiction or fabrication. It may consist of intrinsic or extrinsic hallucinations. This problem occurs due to the probabilistic characteristics of LMs, which prioritize coherence and fluency over factual correctness. Factual contradiction describes cases where LLMs generate responses that are grounded in real-world information but contradict each other. This typically arises due to conflicting training data, entity confusion, or errors in context retention. For example, a model states that \textit{"The capital of Saudi Arabia is Dammam"}, which contradicts the known fact that it is Riyadh. Factual fabrication describes cases where LLMs generate responses that cannot be verified with real-world information. For instance, the model claims that \textit{"Sarah Collins travelled to the planet Mars"} when, in reality, no one travelled to Mars.

\begin{figure*}[t]
    \centering
    \includegraphics[width=\textwidth]{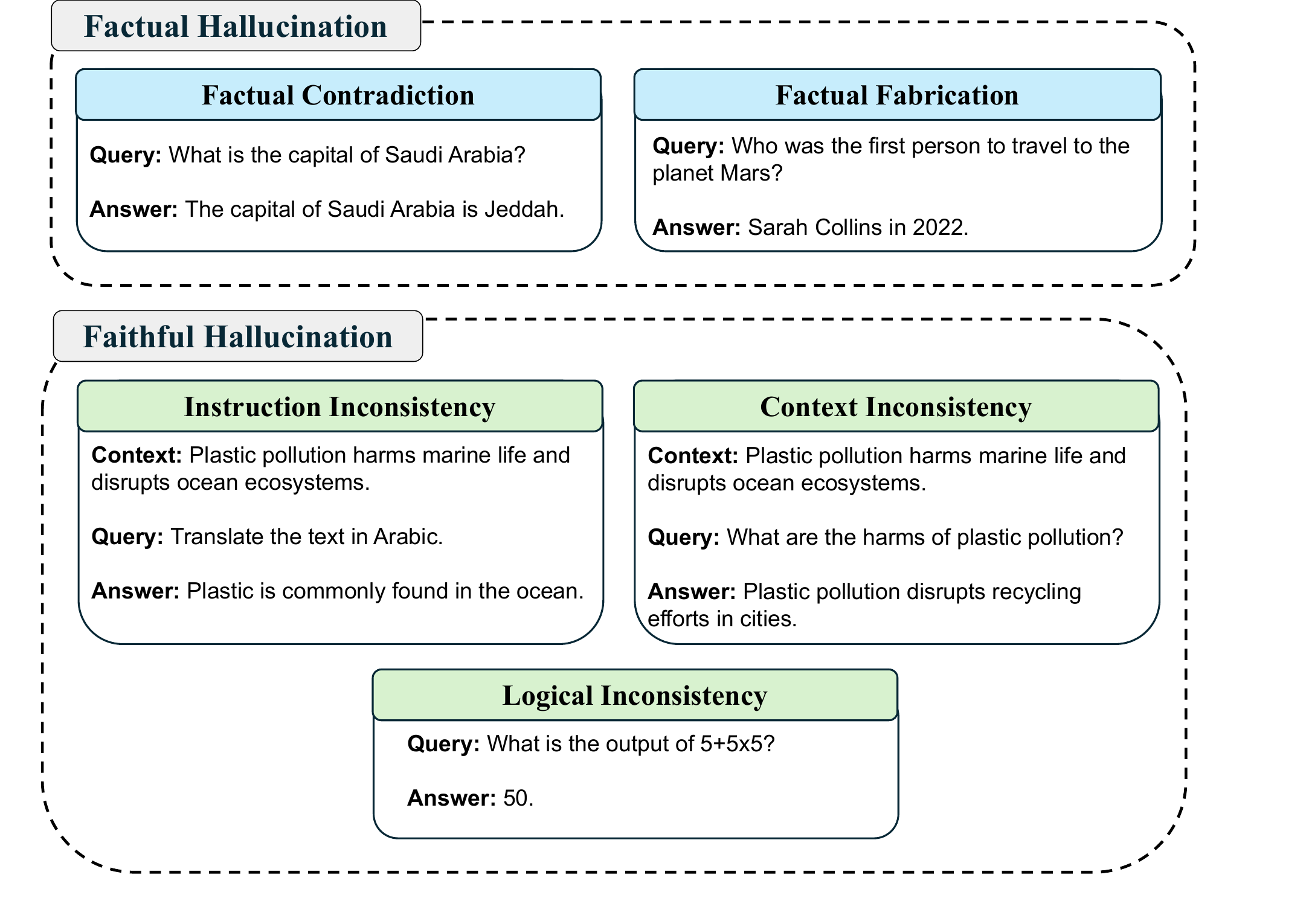}
    \caption{Examples of various hallucination types in LLM responses.}
    \label{fig:factfaith}
\end{figure*}

Faithfulness hallucination occurs when the generated output drifts from the original input or context, violating the user's instructions or violating the logical consistency within the response. This type of hallucination is further categorized into three subtypes: instruction, context, and logical \cite{huang2023survey}. Instruction inconsistency refers to cases where the LLM fails to follow the user's instruction. For example, if a user asks an LLM to summarize a paragraph in one sentence, the LLM generates a full paragraph. Context inconsistency refers to cases where the LLM ignores or alters important facts within the original text. For example, if a given passage states that \textit{"The Mona Lisa was painted by Leonardo da Vinci,"} the model incorrectly states that \textit{"The Mona Lisa was painted in the 17th century"}. Logical inconsistency refers to cases where the LLM's internal reasoning contradicts itself, leading to logically flawed outputs. This may stem from flawed inference or unstable reasoning chains. Figure \ref{fig:factfaith} presents examples of factuality and faithfulness hallucinations in LLMs.

\subsection{Hallucination in NLG Tasks}
\label{sec_NLG}
Hallucination in LLMs appears mainly in NLG tasks that involve generating open-ended text. In contrast, hallucination occurs less frequently in natural language inference (NLI) tasks, such as entailment classification and sentiment analysis, that require the LLM to select from limited and pre-defined options \cite{ji2023survey}. Consequently, they reduce the possibility of fabrication in such tasks compared to NLG tasks. This section describes how hallucination can occur in the main NLG tasks. 

\textit{Machine Translation} involves translating a text from one language to another. Models may provide grammatically accurate translations but may convey new or irrelevant meanings, particularly for low-resource languages or ambiguous inputs \cite{raunak2021curious}.

\textit{Text Summarization} is generating a concise and coherent summary that preserves the most important information from one or more source documents. In the literature, two main methods for text summarization are followed: extractive and abstractive \cite{hahn2000automatic}. Extractive text summarization selects and organizes the most important sentences from the source text to form a summary. Abstractive text summarization, on the other hand, paraphrases the selected sentences to form a summary. Abstractive text summarization is more prone to hallucination than extractive text summarization, since paraphrasing the text may lead to generating false information \cite{maynez2020faithfulness}. 

\textit{Generative QA} is a type of QA system that generates answers to user queries in natural language, rather than selecting a predefined answer from a database or document.  It constructs its response based on the model's comprehension of the inquiry and the provided context. Generative QA systems search for external knowledge from multiple sources to formulate a response. These sources may consist of redundant, complementary, or conflicting information \cite{snyder2024early,shaier2024adaptive}. Therefore, it is highly susceptible to generating hallucinated text.

\textit{Dialogue System} facilitates human-computer interaction through natural conversation. Dialogue systems are broadly categorized into task-oriented and open-domain systems. Task-oriented systems help users complete specific goals, such as booking flights or managing appointments using structured databases \cite{zhang2020recent}. Open-domain systems engage in unrestricted conversations across diverse topics \cite{huang2020challenges}. Open-domain dialogue systems are prone to generating content that is not necessarily present in the input or knowledge source, leading to extrinsic hallucinations. Task-oriented systems, while more constrained, can still hallucinate due to incomplete or misclassified data \cite{ji2023rho,pan2024exploring}

\textit{Data to Text} is a task that entails utilizing structured data, such as a table, as input to generate text that accurately and coherently represents this data as natural language text. The models of this task are prone to hallucination due to the gap between structured data and text \cite{tian2019sticking}. 
\textit{Paraphrasing} is the task of rewording text while preserving its original meaning. The model is required to generate text with semantically equivalent expressions to the source text. Hallucination can occur in this task when the model adds information not present in the source or omits crucial details. This can distort the intended message, resulting in potential misunderstandings \cite{witteveen2019paraphrasing}.

\textit{Code Generation} is an NLG task that generates code snippets based on natural language descriptions. This task is highly prone to hallucination, resulting in code that appears plausible but consists of logical, runtime, or syntax errors \cite{liu2024exploring}.  Such hallucinated code can lead to software malfunctions or security vulnerabilities if not identified and corrected.

\textcolor{black}{\subsection{Impact of Hallucination in Critical Domains}
LLMs Hallucination can pose a considerable challenge, particularly in domains where precise information is essential, such as healthcare, legal counsel, finance, and education. In such domains, LLMs hallucination can directly result into clinical harm, legal liability, and financial loss.} 

\textcolor{black}{\textit{Hallucination in medicine} is defined as the generation of an output that is not supported by authoritative clinical evidence and could alter clinical decisions \cite{kim2025medical}. One of the active research areas in employing LLMs in the medical domain is clinical documentation and consultation summarization \cite{asgari2025framework}. Asgari et al. \cite{asgari2025framework} found that the most common hallucination type in this domain is fabricating the planning section of clinic notes. However, detecting hallucination in this domain is challenging due to the ambiguity of the terms, which leads to diverse errors \cite{kim2025medical}. Therefore, detecting hallucination in the medical domain depends highly on the level of domain expertise and the level of detail in the prompt provided to the detection model. Domain experts are more likely to identify subtle inaccuracies in clinical terminology and reasoning, whereas non-experts may struggle to discern these errors, thereby increasing the risk of misinterpretation \cite{kim2025medical}. Accordingly, transparency, interpretability, and trustworthiness in medical AI are a necessity for Healthcare professionals to understand how and why an LLM arrived at its conclusions and thereby improve hallucination detection and mitigation in the medical domain.}

\textcolor{black}{\textit{Hallucination in finance} raises a critical risk, since hallucinated numbers, mis-summaries of financial statements, or fabricated market events can drive misallocation of capital \cite{kang2023deficiency}. Various specialized LLMs, such as FinBERT \cite{yang2020finbert} and BloombergGPT \cite{wu2023bloomberggpt} have been developed to generate financial-related content. However, hallucination presents a major challenge to deploy such LLMs in real-world applications \cite{kang2023deficiency}. To mitigate hallucination in the financial domain, Kang and Liu \cite{kang2023deficiency} demonstrated the effectiveness of RAG with prompt-based tools in learning to generate accurate content. Moreover, Roychowdhury \cite{roychowdhury2024journey} demonstrated the effectiveness of prototyping, scaling, and LLM evolution using human feedback to minimize hallucination in the financial domain.}

\textcolor{black}{\textit{Hallucination in law} poses a significant risk due to the authoritative tone and persuasive fluency of LLM-generated legal text, which may obscure factual or doctrinal inaccuracies. Legal hallucination refers to the generation of incorrect, fabricated, or outdated legal information, such as non-existent case law, misquoted statutes, or erroneous legal reasoning, that is not supported by valid legal sources. Such hallucinations are particularly dangerous because legal decisions often rely on precise interpretations of jurisdiction-specific laws, precedents, and procedural rules \cite{dahl2024large}. Detecting hallucination in the legal domain is particularly challenging due to the complexity and evolving nature of legal systems in different countries. Additionally, legal validity depends on temporal factors, jurisdictional scope, and contextual interpretation, all of which are difficult for LLMs to model reliably \cite{dahl2024large}. As a result, hallucination detection in legal applications requires access to up-to-date legal corpora, robust citation verification mechanisms, and domain-aware evaluation criteria that go beyond surface-level factual consistency.}

\textcolor{black}{Beyond model-level evaluation, hallucination detection and mitigation in critical domains must be integrated into practical system workflows. In healthcare, LLM outputs should be deployed as decision-support tools, with hallucination detection techniques serving as real-time safeguards that flag uncertain or unsupported content. Such human-in-the-loop designs align with clinical governance requirements and reduce the risk of automation bias. Similarly, in financial applications, hallucination mitigation must be coupled with auditability, confidence estimation, and traceability. This is to ensure that generated analyses and summaries can be verified against authoritative sources before influencing high-stakes decisions. Moreover, in the legal domain, LLMs must be integrated into human-in-the-loop workflows, where generated outputs serve as assistive drafts rather than final legal judgments. Human oversight by qualified legal professionals is essential to validate citations, reasoning chains, and conclusions before use in real-world settings. Moreover, regulatory and ethical considerations, including professional responsibility rules, data privacy laws, and accountability frameworks, necessitate transparency, traceability, and auditability in legal AI systems. Regulatory considerations further necessitate transparency and accountability, as hallucinated outputs may lead to legal liability or regulatory violations. Consequently, effective hallucination detection and mitigation in real-world systems requires a combination of technical safeguards, domain-aware human oversight, and compliance-aware deployment strategies.}

\section{Hallucination Causes}
\label{sec:sources}

\begin{figure*}[t]
    \centering
    \includegraphics[width=\textwidth]{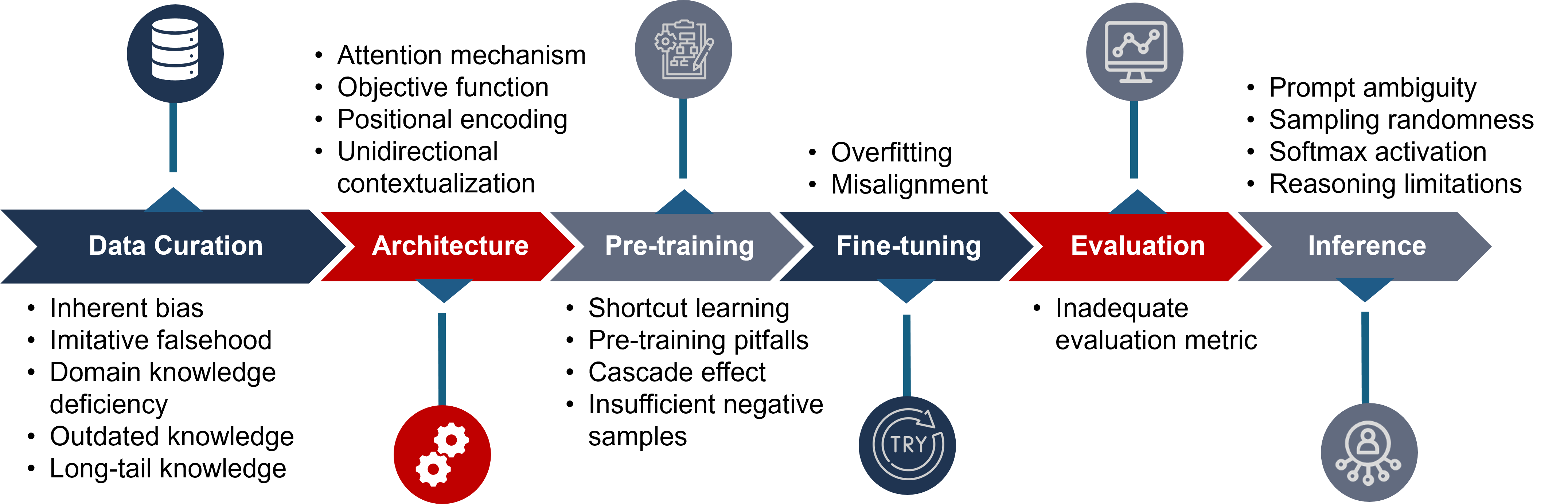}
    \caption{The main causes of hallucination at different stages of the LLM development pipeline.}
    
    \label{fig:Casues}
\end{figure*}

Several reasons can make LLMs hallucinate when responding to the user query. These causes can be linked to the LLM development lifecycle, starting from pre-training to the inference pathway. Figure \ref{fig:Casues} illustrates the main stages of developing LLMs and the causes of hallucination associated with each stage. This section discusses hallucination sources at each stage of the LLM development process. 

\subsection{Data Curation}
The initial stage of LLM development is data collection and preparation. Large-scale and diverse datasets are usually collected from different sources, such as books, websites, social media, and scientific articles. These datasets encompass diverse domains, languages, and contexts. Once collected, the data undergoes extensive filtering and cleaning to remove duplicates, irrelevant, and noisy content.  

Although LLMs' capabilities are significantly improved by scaling up pre-training data \cite{kaplan2020scaling}, scaling introduces persistent challenges in maintaining data quality \cite{lin2022truthfulqa}. The utilized data for training LLMs is a significant source of \textit{bias}, which LLMs may unintentionally acquire and propagate \cite{ferrara2023fairness}. Online content, in particular, can reflect societal imbalances in the representation of gender, race, nationality, and other demographic factors \cite{bolukbasi2016man}. These social biases are inherently linked to hallucinations. Additionally, LLMs may exhibit memorization tendencies, particularly with frequently occurring data points, despite deduplication efforts during the data preparation stage. As a result, the model might over-represent high-frequency words, phrases, and concepts from the training data, which creates an imbalance in the generated outputs. Consequently, they exhibit a bias towards over-represented training data, which results in a hallucinated output that diverges from the desired content \cite{lee2022deduplicating}. 


Another source of data-induced hallucination is \textit{imitative falsehoods} \cite{lin2022truthfulqa}. It arises when a model learns and reproduces false or misleading information embedded in its training data, often originating from misconceptions or misinformation. LLMs are designed to mimic the patterns in their training distribution, which can accidentally amplify common misconceptions. For instance, if an LLM encounters incorrect attribution of the creation of the light bulb exclusively to Thomas Edison, it is prone to reproduce this error upon inquiry. Although Edison played a crucial role in the development and commercialization of the light bulb, he was not its original inventor. However, due to the prevalence of simplistic historical narratives, the model may confidently present this misunderstanding as fact, perpetuating an imitative deception.

Furthermore, \textit{knowledge conflict} is another source of hallucination, wherein the model is trained using different sources that provide contradicting information regarding the same subject. These inconsistencies can lead to outputs that reflect conflicting viewpoints or factual errors \cite{wang2023resolving}.
\textit{Domain knowledge deficiency} can also make LLMs hallucinate. Despite the impressive performance of LLMs in zero-shot scenarios, they struggle with tasks that require specialized reasoning or access to confidential data \cite{topol2019deep}. In fields such as healthcare, precision and accuracy of information are essential. The lack of domain-specific training data can lead to hallucinations, which often appear as factual inaccuracies \cite{huang2023survey}.

Another source of hallucination in LLMs is \textit{outdated factual knowledge}. 
Once LLMs are trained, their internal parametric knowledge remains fixed and does not reflect subsequent changes in real-world facts. Therefore, LLMs often generate fabricated facts or responses that were once accurate but are now outdated when faced with questions outside their training time-frame \cite{mousavi2024your}. This temporal misalignment leads to hallucinated content, which compromises the factual reliability of LLM outputs. 

Additionally, LLMs are especially prone to hallucinations when dealing with \textit{long-tail knowledge} that appears infrequently in the training data \cite{kandpal2023large}. LLMs are trained to recognize patterns in text based on how often words and phrases appear together. As a result, they tend to perform well on common or frequently discussed topics. However, when it comes to rare or obscure entities that are not well represented in the training data, LLMs are more likely to produce inaccurate or entirely fabricated responses \cite{kandpal2023large}.


\subsection{Model Architecture}


While architectural design primarily aims to enhance learning capabilities and efficiency, some learning strategies or LLM's architecture components can inadvertently contribute to hallucinations. The sources of the hallucination related to the model architecture involve attention mechanism, objective function,  positional encoding, and unidirectional contextualization. 
Attention enables LLMs to dynamically focus on different segments of the input when generating an output \cite{vaswani2017attention}. Self-attention helps capture long-range dependencies within the same sequence, while cross-attention enables conditioning on external context, such as retrieved documents. However, the soft attention mechanisms in capturing long sequences can result in hallucination \cite{liu2024survey}. As the sequence length increases, the attention weights may become more diffuse, leading the model to distribute the focus between less relevant tokens, which can result in degraded reasoning or factual inaccuracies \cite{liu2024exposing}.

The \textit{objective function} used during model training can influence the likelihood of LLM hallucination \cite{welleck2019neural}. Most models use maximum likelihood estimation (MLE), which encourages generating the most probable token at each step. However, this method does not explicitly penalize factual inconsistencies, which can lead to hallucinations, especially when a model confidently fills in missing information based on statistical likelihood \cite{welleck2019neural}.
\textit{Positional encoding} also plays a fundamental role in LLMs, as transformers lack intrinsic awareness of tokens order.
LLMs augment input sequences with positional encodings; either using fixed sinusoidal functions, as in the original transformer, or learned position embeddings, as adopted in models like GPT-3 and T5. These methods enable LLMs to learn tokens order and maintain context for moderate-length texts. However, as input sequences become longer, the effectiveness of these positional representations tends to deteriorate \cite{banerjee2024llms}. Consequently, the model may not longer reliably track relative positions, which results in misinterpreting which tokens are related or contextually relevant. As a result, hallucinations may emerge when the model misinterprets contextual relationships due to position-tracking limitations.

Lastly, \textit{unidirectional contextualization} can cause hallucination. Autoregressive LLMs, such as GPT, process text unidirectionally, in a left-to-right fashion. This behaviour inherently limits their capacity to comprehensively capture and integrate contextual information from both preceding and subsequent tokens. Thus, driving the model to depend primarily on local patterns. Consequently, when the model is faced with ambiguous or incomplete input, it may infer or fabricate content to maintain coherence, thereby introducing hallucinations \cite{huang2023survey}.

\subsection{Model pre-training}
The following stage of the LLM development process is pre-training the model. This stage utilizes massive data to train the model to learn general language representation in an unsupervised manner. The model is generally trained with a language modeling objective, where the model learns to predict the next token in a sequence given its preceding context. While pre-training significantly improves LM performance and generalization, some strategies employed during this phase may lead to hallucination during inference, such as shortcut learning, teacher forcing learning strategy, and a lack of sufficient negative examples.

\textit{Shortcut learning} is a phenomenon where a model tends to learn superficial, non-robust patterns of the data, rather than robust features for making predictions \cite{bihani2024learning}. This over-reliance on certain characteristics or biases can lead to inadequate generalization in out-of-distribution contexts. Some studies indicate that LLMs often exploit shortcuts derived from statistical indicators, such as the word "not" \cite{niven2019probing}, specific keywords \cite{du2021towards}, and cues associated with linguistic variations \cite{nguyen2021learning}, to formulate predictions. Consequently, LLMs often generate dependable results with independent and identically distributed samples but may exhibit hallucinations with out-of-distribution data. 
Another cause of hallucinations is the \textit{teacher forcing learning strategy} \cite{kang2020improved}. In the teacher forcing MLE setup, the model learns to predict the next word in a sequence based on a flawless context. During inference, the model predicts each following token based on its previously generated tokens, rather than relying on ground truth inputs. This discrepancy, known as exposure bias, can cause the model to hallucinate if an early token is incorrect or contextually inappropriate \cite{wang2020exposure}. This deviation is further exacerbated by the cascade effect, where an early mistake leads to a chain reaction of subsequent errors, which can compound errors in a "snowball effect." This can be attributed to the lack of corrective feedback when the model generates an incorrect token \cite{zhang2023language}. This issue arises particularly in high-entropy segments where the model's confidence is low, increasing the likelihood of hallucinations.

Moreover, the \textit{lack of sufficient negative examples} during training can weaken the model’s ability to distinguish between fact and fiction \cite{hamdan2025much}. While LLMs rapidly attain exceptional performance on benchmark tasks, they often struggle with simple challenge instances and underperform in real-world situations \cite{kiela2021dynabench}. Without exposure to diverse incorrect or misleading examples during training, models can fail to recognize and correct common misconceptions \cite{kiela2021dynabench,nie2020adversarial}. During a critical phase of training, each negative example can improve model accuracy up to ten times more than a positive one, as it helps the model sharply reduce the likelihood of plausible but false answers \cite{hamdan2025much}.


\subsection{LLMs Fine-Tuning}

LLMs are usually fine-tuned after the pre-training stage on more specialized datasets related to downstream tasks, such as healthcare reports summarization, QA, and stance detection. Supervised fine-tuning is an iterative process that includes re-training the model's parameters partially until the desired capabilities are met. In parallel with supervised fine-tuning, reinforcement learning from human feedback (RLHF) has emerged as a key strategy for aligning model output with human preferences.  RLHF generally employs a preference model that receives rewards from human evaluators who judge its generated responses based on some criteria, such as factual accuracy and relevance \cite{christiano2017deep}. To conform to human preferences, RLHF guides the LLM to produce outputs that maximize the reward given by the trained preference model, usually via a reinforcement learning (RL) \cite{schulman2017proximal}. This approach has proven effective in aligning LLMs with human intent and improving output quality. However, both supervised fine-tuning and RLHF introduce some risks that can lead to hallucinations during inference.

One reason for hallucination at this stage is \textit{overfitting} on task-specific data \cite{pan2025towards}. When models are fine-tuned exclusively on narrow or domain-specific datasets, they usually become overly sensitive to the patterns and biases present in that data. Domain-specific fine-tuning may constrain the model's generalization and increase the likelihood of generating deterministic and biased solutions that misrepresent the original data distribution \cite{gekhman2024does}. If this model is later exposed to out-of-distribution prompts, it may attempt to generate answers beyond its learned domain, which increases the risk of hallucination.

Another source of hallucination at this stage is the \textit{misalignment} between the model’s internal capabilities and the expectations encoded in the alignment data  \cite{huang2023survey}. Alignment refers to the process of ensuring that the model’s outputs are aligned with human preferences. Although alignment significantly improves the quality of LLM responses, it also increases the risk of hallucination \cite{huang2023survey}. This risk arises when there is a mismatch between the model’s intrinsic capabilities and the alignment data’s expectations. One of these misalignments is \textit{capability misalignment}, which occurs when alignment training encourages the model to provide definitive answers even when it lacks sufficient knowledge \cite{huang2023survey}. Although RLHF encourages the model to generate responses that meet human preferences, it may prioritize coherence and confidence over factuality, which leads to hallucinated responses. Another misalignment category is \textit{belief misalignment}, where disparity occurs between the model's internal beliefs or knowledge learned from pre-training and its output after alignment \cite{sharma2023towards}. This issue often occurs alongside sycophantic behavior, a tendency for the model to generate responses that evaluators will approve of, regardless of whether those responses are accurate or not \cite{sharma2023towards}.


\subsection{LLMs Evaluation}
This stage involves evaluating the model's generation accuracy and coherence using benchmark datasets for downstream tasks, such as QA and summarization. This assessment is accomplished by both automatic and human evaluation. One of the automatic metrics used to evaluate the LLM on intrinsic language tasks is perplexity. It evaluates the likelihood of a sequence of words given the model’s learned probabilities. 
Human evaluators assess model responses based on coherence, fluency, faithfulness, and factuality. Ensuring factual alignment is crucial for mitigating hallucination, as models may generate fluent yet misleading responses. Therefore, the outcomes of these evaluations are then used to improve the model's factual consistency and coherence. 

\textit{Inadequate evaluation metrics} are one of the main reasons for undetected hallucination in this stage \cite{maynez2020faithfulness}. Automatic metrics, such as ROUGE \cite{lin2004rouge}, BertScore \cite{zhang2019bertscore}, and BLEU \cite{papineni2002bleu}, are usually used for evaluating LLMs. However, these metrics often fail to assess the factuality and faithfulness of the generated text. This can lead to models that perform well on such metrics but hallucinate during deployment \cite{maynez2020faithfulness}.

\subsection{Inference}
After deployment, LLMs are ready for the inference stage to generate responses to user queries in real-time. During this stage, the model leverages its pre-trained knowledge and fine-tuning to deliver relevant output to the users' inputs. However, despite optimization and alignment efforts, deployed models remain vulnerable to hallucinations, especially when encountering ambiguous inputs, randomness in sampling, architectural limitations, or reasoning challenges.




One common source of hallucination in the inference stage is \textit{ambiguous input prompts}. Ambiguity is a natural part of language, which involves multiple alternative meanings and contextual relationships for the language unit. Prompts that are vague, ambiguous, or prone to speculation are often a source of hallucinations \cite{rawte2023troubling}. A lack of specificity in user queries encourages LLM to rely on its own training data and experience rather than addressing the user's request. For instance, given the vague prompt, \textit{“Explain recent breakthroughs in energy,”} the LLM might respond with a fabricated claim such as \textit{“One recent breakthrough is ‘quantum solar cells,’ developed by Dr. Sarah Lin in 2023, which convert sunlight into energy with 95\% efficiency, a revolutionary improvement over traditional solar cells.”} This vague prompt led the LLM to invent a non-existent breakthrough and research, resulting in a hallucination.

Another source of LLM hallucination at this stage is \textit{inherent sampling randomness} \cite{dziri2021neural}. During text generation, the model selects the next word based on a probability distribution over possible tokens. While deterministic decoding strategies (e.g., greedy decoding) favor high-probability tokens and minimize hallucination risk, they often lead to repetitive or uninspired responses. This is a phenomenon known as the likelihood trap \cite{meister2020if}. In contrast, stochastic decoding methods such as top-k or nucleus sampling introduce creativity. However, these methods also increase the chance of selecting low-probability tokens that diverge from factual or contextual correctness \cite{dziri2021neural}. As randomness increases, the model is more likely to draw from the tail of the distribution, leading to vivid but potentially inaccurate generations.

\textit{SoftMax activation} can also cause hallucinations during inference \cite{yang2017breaking}.  
LLMs often use a softMax function to calculate word prediction probability to predict the likelihood of each next word in the sequence. SoftMax is optimized for contexts where there is one dominant next word. However, in rich contexts with multiple potential meanings, the desired distribution might have multiple peaks corresponding to different word choices. SoftMax struggles to handle these situations, as it cannot easily represent multiple equally relevant words. This limitation, known as the softMax bottleneck \cite{yang2017breaking}, occurs when the model’s output layer cannot adequately assign high probabilities to multiple diverse yet equally relevant words, restricting its ability to represent them equally well.


Finally, \textit{reasoning limitations} pose a substantial challenge for factuality in LLM outputs \cite{zheng2023does}.
LLMs often fail to produce accurate responses in scenarios that require multi-hop reasoning or logical deduction. Complex reasoning tasks involve linking multiple pieces of information or making logical inferences. This can extend beyond simple recall and requires a structured understanding of relationships within the data. In such multi-hop QA scenarios, the model needs to connect different pieces of information across multiple steps \cite{zheng2023does}. For instance, if a question requires the use of one fact to understand another, then the LLM must perform the necessary reasoning that bridges these two steps to arrive at a correct answer. If the model fails to link the facts properly, it will usually hallucinate when generating responses.


\section{Hallucination Detection}
\label{sec_detection}
\begin{figure*}[h]
    \centering
    \includegraphics[width=0.8\linewidth]{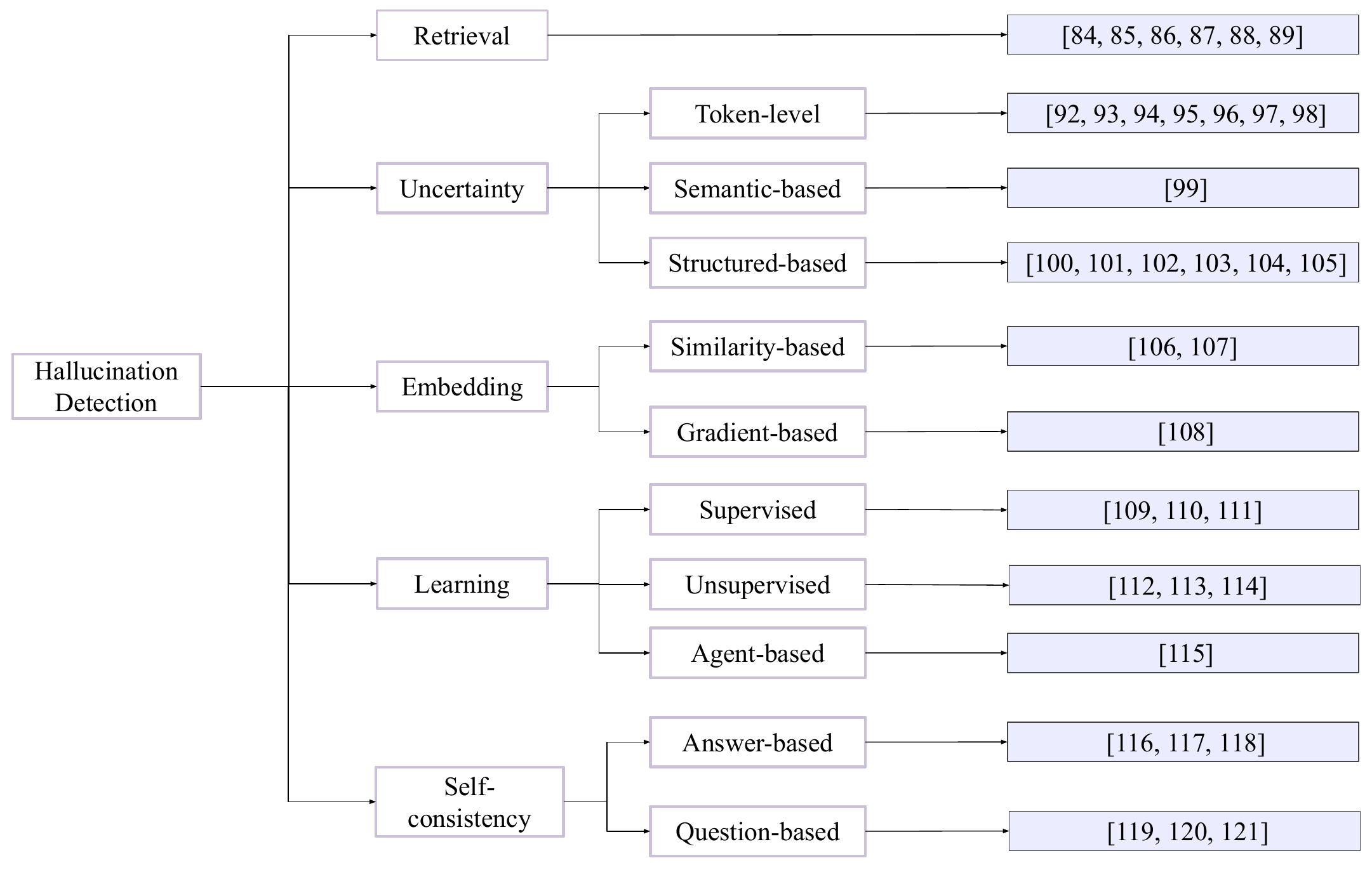}
\caption{The taxonomy of hallucination detection methods.}
\label{fig:detection}

\end{figure*}

Hallucination detection involves identifying instances in LLM outputs that are inaccurate, nonsensical, or inconsistent with the input or context. Unlike traditional fact verification, which primarily verifies claims against external sources, hallucination detection involves a more comprehensive analysis. It often requires analyzing \textcolor{black}{factuality and faithfulness} within the model's responses. In this survey, we categorize hallucination detection methods present in the literature into retrieval, uncertainty, embedding, learning, and self-consistency-based techniques. The proposed taxonomy is illustrated in Figure \ref{fig:detection}. Table \ref{tab:detection-comparison} summarizes the hallucination detection studies. The results are mainly reported using the AUROC metric, which tests the model's ability to distinguish between the hallucination and non-hallucination classes across all possible probability thresholds.



\subsection{Retrieval-Based Detection}
Retrieval-based hallucination detection methods compare an LLM's output with trusted external knowledge sources, such as databases and encyclopedias. These methods aim to ensure factual consistency and reduce the risk of unsupported content by grounding the output in verifiable information. RAG is one of the most widely used retrieval-based techniques for hallucination detection \cite{rakin2024leveraging, sriramanan2024llm,zhu2024halueval}. It combines retrieval systems with generative models to fetch relevant documents and compare their content with LLMs-generated outputs \cite{niu2024ragtruth}. RAG can be followed by fact-checking models to evaluate the correctness of statements \cite{lewis2020retrieval,niu2024ragtruth}. 

\textcolor{black}{Ajmal et al. \cite{ajmal2025evaluating} detected hallucinations by assessing the semantic and factual consistency of generated answers against the source. The approach uses contextual similarity analysis and cross-referencing to identify discrepancies, such as incorrect entities, factual contradictions, or logically inconsistent statements. The detected discrepancies are quantified through a scoring mechanism that reflects the magnitude of hallucination. Similarly, Paudel et al. \cite{paudel2025hallucinot} introduced a hallucination detector for enterprise applications by verifying LLM outputs against the provided context, which is the retrieved document and common knowledge, which is the LLM's parametric knowledge. It uses a modular detector (HDM-2) that checks whether generated statements contradict the provided context or conflict with common knowledge. Detection is performed at both the response level and span level, producing hallucination scores and fine-grained annotations.} 

In order to improve retrieval-based techniques, Wang et al. \cite{wang2023hallucination} used a Bayesian sequential estimation. Instead of retrieving a pre-defined number of documents, one document is retrieved at a time to assess each subclaim's veracity. After each retrieval, the Bayesian sequential analysis decides whether to continue retrieving documents or stop. This dynamic stop-or-continue strategy optimizes the balance between retrieval costs and accurate hallucination detection, which leads to improved efficiency and precision. Building on the need for more granular analysis, Mishra et al. \cite{mishra2024fine} proposed FAVA, a retrieval-augmented model trained on synthetic data to detect various types of hallucination using span-level detection. FAVA outperformed baseline models, including GPT-4, in both span-level and binary hallucination detection and in factual editing tasks. KnowHalu is another improvement over retrieval-based hallucination detection \cite{zhang2024knowhalu}. It first detects non-fabrication hallucination, then performs a multi-form factual check through step-wise reasoning, query decomposition, and knowledge retrieval. By aggregating judgments across multiple knowledge forms, KnowHalu achieved significant improvements of 15.7\% and 5.5\% in QA and summarization tasks, respectively. \textcolor{black}{In another study, JointCQ \cite{xu2025jointcq} detected hallucination by extracting factual claims and generating targeted search queries from a QA pair. These queries are then used to retrieve external evidence. A verifier model checks whether each claim is supported, contradicted, or unverifiable based on the retrieved contexts. Using HalluQA, it achieved an accuracy of 80.58\% and an F1-score of 83.05\%.}


\subsection{Uncertainty-Based Detection}
\begin{figure*}[t]
    \centering
    \includegraphics[width=\linewidth]{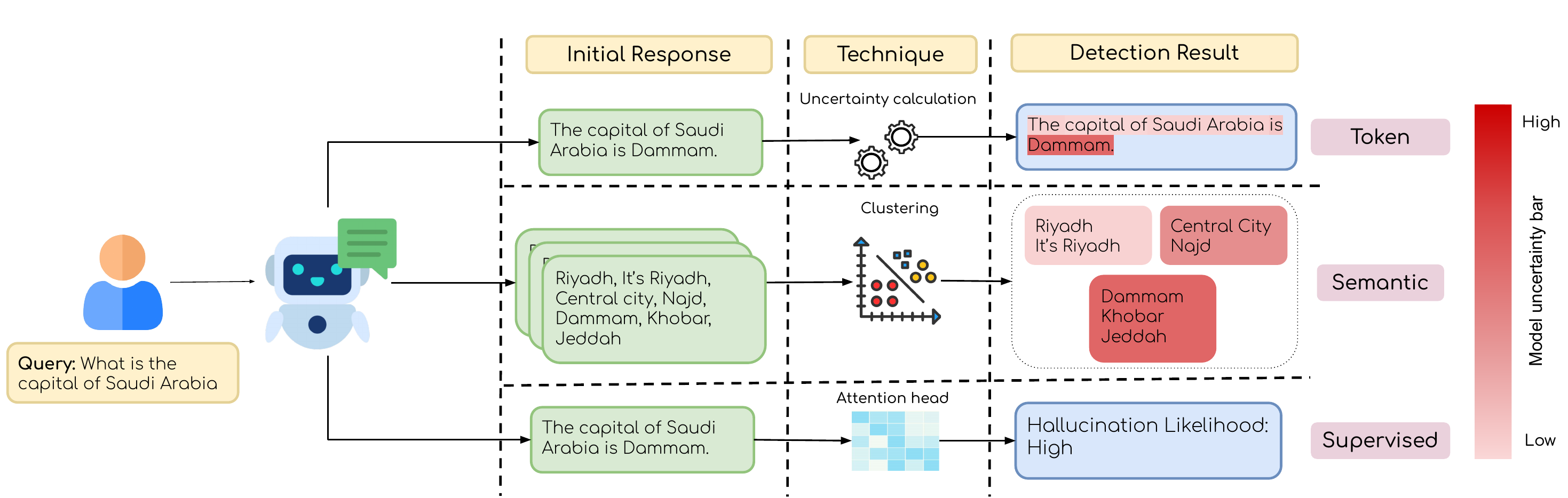}
    \caption{Examples of hallucination detection techniques across three uncertainty-based approaches: token-level uncertainty calculation, semantic clustering, and \textcolor{black}{structured} detection via attention heads.}
    
    \label{fig:uncertainty}
\end{figure*}

Probabilistic and uncertainty-based hallucination detection methods flag low-confidence outputs as potential hallucinations without requiring external knowledge retrieval. These methods exploit the inherent uncertainty in the model predictions to identify hallucinated content. The main hypothesis in uncertainty estimation is that high uncertainty indicates that the model is guessing rather than relying on learned patterns \cite{xiong2023can,huang2023look}. Uncertainty-based hallucination detection techniques can be classified into token-, semantic-, and structure-based methods. Examples of the differences between these techniques are shown in Figure \ref{fig:uncertainty}.

\textbf{Token-based Approaches.} Several methods have been used to quantify token-level uncertainty in LLMs to detect hallucination. Guerreiro et al. \cite{guerreiro2023looking} detected LLM hallucination in machine translation tasks using sequence log-probability. This approach measures the confidence of the model by calculating the normalized log probability of the generated tokens. In another study \cite{yang2023improving}, the log-based uncertainty measure was used to develop an uncertainty-aware framework for hallucination detection. The method relies on token probability scores derived from LLM logit outputs to measure uncertainty. Instead of setting a strict threshold, uncertainty is introduced as an intermediary variable to be used adaptively. Accordingly, the model implicitly incorporates uncertainty in its decision-making. Although token-level uncertainty estimation achieved promising results, this technique is calculated based on lexical variety. Therefore, responses that give the same meaning but use different words will not be treated as uncertain. Zhang et al. \cite{zhang2023enhancing} enhanced uncertainty-based hallucination detection by improving token probability estimation, reducing overconfidence, and incorporating focus mechanisms. This approach focuses on the most informative keywords, unreliable tokens in historical contexts, and specific token properties. Instead of relying on raw probability, Dasgupta et al. \cite{dasgupta2025hallushift} proposed HalluShift, which detects hallucinations by measuring distribution shifts in internal hidden states and attention layers. 

\textcolor{black}{More recent work improves upon this idea by learning to model token-level uncertainty from internal representations rather than relying only on raw probability values. Instead of treating all tokens equally, these methods train auxiliary models to identify which token-level signals, such as hidden states or probability distributions, are most indicative of hallucination.  Shelmanov et al. \cite{shelmanov2025head} proposed pre-trained uncertainty quantification heads to predict claim-level hallucination by leveraging attention maps and token probability features within the LLM. These transformer-based heads are trained on annotated hallucination data, which outperformed both classical unsupervised and prior supervised uncertainty methods, with strong generalization across domains and languages. Similarly, Niu et al. \cite{niu2025robust} proposed a multiple instance learning framework for hallucination detection that adaptively selects salient token embeddings that are most indicative of factual inaccuracy. The selection is guided by uncertainty metrics such as token-level and sentence-level entropy. By combining these uncertainty features with deep representation learning, this framework achieved robust and generalizable hallucination detection, outperforming prior supervised and unsupervised uncertainty-based methods.} \textcolor{black}{HaluNet \cite{tong2025halunet} detected hallucinations using internal uncertainty signals extracted in a single forward pass. It combines token log-likelihoods (confidence), entropy (distributional uncertainty), and hidden-state embeddings (semantic uncertainty) through a lightweight multi-branch neural architecture. These uncertainty signals are fused with attention-based weighting to learn how different types of uncertainty correlate with hallucination. Unlike self-consistency methods, it does not require multiple generations, making it efficient and suitable for real-time QA systems.}

\textbf{Semantic-based Approaches.} Farquhar et al. \cite{farquhar2024detecting} proposed a semantic entropy measure to detect hallucinatory content. Semantic entropy is computed over the sentence's meaning rather than relying on the words' distribution. This technique first generates multiple responses to the same prompt. These responses are then clustered based on semantic similarity. Moreover, entropy is computed across these semantic clusters. A high semantic entropy indicates that the model’s answers vary in meaning, indicating a high likelihood of hallucination, while a low semantic entropy suggests a consistent meaning across generations, which means greater reliability.

\textcolor{black}{\textbf{Structured-based Approaches.} Recent studies proposed structured uncertainty detection, which detects hallucination by modeling the logical or semantic dependencies between claims as a graph or tree and propagating uncertainty across the connected units.} Hou et al. \cite{hou2024probabilistic} proposed detecting hallucination by belief tree propagation (BTPROP). Instead of relying solely on token-level uncertainty, BTPROP recursively decomposes a statement into logically related subclaims and builds a tree of beliefs. A hidden Markov tree is then used to integrate the model's noisy confidence scores and logical relationships between claims. This approach robustly propagates uncertainty and corrects miscalibrated beliefs. Expanding on these directions, Chen et al. \cite{chen2025enhancing} introduced a graph-enhanced uncertainty modeling approach to capture the relations among entity tokens and sentences. The authors also proposed a graph-based uncertainty calibration technique that accounts for the possibility of phrase conflicts with neighbors in the semantic graph when calculating uncertainty. This method achieved a notable improvement in passage-level hallucination detection, increasing the Spearman correlation by 19.78\%.

Beyond semantic graphs, recent work explores structural signals directly derived from internal attention patterns. Lookback Lens \cite{chuang2024lookback} is an unsupervised approach that analyzes attention for hallucination detection. By measuring the lookback ratio, which quantifies how much the model attends to prior context versus its own outputs, this approach identifies contextual hallucinations without requiring labeled data. This method achieved competitive performance with supervised detectors like RIPA and supports real-time mitigation during decoding, though it incurs higher inference time and relies on effective sampling. LapEigvals \cite{binkowski2025hallucination} modeled attention maps using a Laplacian graph and spectral features. It achieved superior results, with an AUC-ROC of 88.9\% on the TriviaQA dataset. \textcolor{black}{Likewise, HalluZig \cite{samaga2026halluzig} detected hallucinations by training a supervised classifier on topological features extracted from attention dynamics. It models layer-wise attention matrices as graphs and applies zigzag persistence from topological data analysis to capture how attention structures evolve across layers. Hallucinations are identified when these signatures indicate unstable or short-lived attention patterns associated with flawed reasoning. Similarly, TOHA \cite{bazarova2025hallucination} detected hallucinations by analyzing the topological structure of attention maps inside the LLM. It builds attention graphs that links between the prompt and response tokens. Then, they compute a topological divergence score (MTop-DivG) that measures how structurally different the response is from the prompt. The method identifies a small set of hallucination-aware attention heads whose divergence consistently separates hallucinated from grounded outputs. The final detection is performed by averaging the divergence scores across these heads, which makes the method computationally efficient.}

\subsection{Embedding-Based Detection}
Embedding-based hallucination detection techniques measure the semantic similarity between input, output, and external references using vector embeddings. These approaches assume that the output of a faithful model should be semantically close to its sources in a shared embedding space. These approaches can be classified into similarity-, gradient-, and spectral-based approaches. 

\textbf{Similarity-based Approaches.}
Dale et al. \cite{dale2023detecting} used cross-lingual semantic comparison between source and translation to detect hallucinations. They evaluated embedding-based similarity models such as LaBSE and LASER, as well as an XNLI entailment model that measures bidirectional semantic consistency. Low semantic alignment between source and translation indicates semantic detachment, which corresponds to hallucination. Among external methods, LaBSE performed substantially better than LASER. Nonkes et al. \cite{nonkes2024leveraging} extended embedding-based approaches by constructing a semantic similarity graph where each generated sentence is a node and the edges connect semantically close outputs. A Graph Attention Network then performs message passing to learn structural patterns in the embedding space, under the hypothesis that hallucinated generations occupy distinct regions of the latent space.


\textbf{Gradient-based Approaches.} Hu et al. \cite{hu2024embedding} extended embedding-based hallucination detection by incorporating gradient information, which captures how sensitively the model’s output responds to its input. Their method characterizes the disparity between conditional and unconditional outputs by a Taylor series expansion, capturing both embedding changes and gradient-based uncertainty signals. A multi-layer perception classifier trained on these features demonstrated SOTA performance across hallucination detection benchmarks.


\subsection{Learning-Based Detection}
Learning-based detection approaches leverage trained models to classify LLM outputs as hallucinated or factual. These models are typically trained on annotated datasets or proxy labels and aim to capture patterns indicative of hallucination beyond rule-based or similarity-based techniques. The learning-from-data feature enables these methods to generalize to diverse hallucination types across domains and tasks.


\textbf{Supervised-based Methods.}
Choi et al. \cite{choi2023kcts} introduced RIPA, a token-level hallucination detector trained on synthetic examples generated through knowledge shuffle and partial hallucination strategies. In the partial hallucination approach, only certain words or phrases in a sentence that are factually correct are replaced with false information. This teaches the model to spot hallucinations that appear in specific parts of a sentence rather than across the whole output. RIPA then identifies both the onset, where the first hallucinated token appears, and the span, the consecutive set of tokens that make up the hallucinated content. In a related direction, Zhang et al. \cite{zhang2024prompt} proposed PRISM, which leverages prompt-guided internal hidden states as features for a supervised hallucination detector. By crafting prompts that enhance the truthfulness structure in the LLM’s internal representations, PRISM achieves good domain generalization, outperforming previous internal-state and token-probability-based baselines in both accuracy and AUROC. In another study, \textcolor{black}{HaluGNN \cite{kong2025halugnn} formulates hallucination detection as a graph classification problem using a supervised GNN. Each QA instance is converted into a weighted directed graph where nodes are token hidden states and edges are attention weights, preserving both semantic content and token relationships. A graph neural network is trained on a small labeled set to distinguish factual vs hallucinated responses.}

\textbf{Unsupervised and Weakly-based Methods.}
\textcolor{black}{Park et al. \cite{park2025steer} introduced a truthfulness separator vector (TSV) that is added to internal LLM activations during inference to reshape the embedding space. TSV aims to increase separability between truthful and hallucinated representations without fine-tuning the model. It is trained using a small labeled set and pseudo-labeled unlabeled generations via optimal transport alignment and confidence filtering.} \textcolor{black}{In another study, Yamada and Arase \cite{yamada2025light} detected hallucination by training a lightweight encoder-based model with contrastive (triplet) learning. The learning objective pulls faithful generations closer to the input in embedding space while pushing hallucinated generations farther away. This contrastive objective is combined with a standard classification loss to improve separability between hallucinated and non-hallucinated outputs. At inference time, the model simply compares the input and generated text to predict whether hallucination is present, without retrieval or external knowledge sources. In another study, contextual embeddings were utilized to detect hallucinations in a real-time, unsupervised manner \cite{su2024unsupervised}. The authors proposed using the internal hidden states of the LLM during inference. They introduced MIND, a lightweight hallucination classifier trained using pseudo-labeled data automatically constructed from Wikipedia. The method records contextualized token embeddings produced at each decoding step and feeds them into an MLP to detect hallucination. The best results were obtained using the last token's embedding from the last transformer layer.}

\textbf{Agent-based Methods.}
Cheng et al. \cite{cheng2024small} proposed HaluAgent, an autonomous hallucination detection agent built on small open-source LLMs. HaluAgent integrates a multi-stage detection pipeline. The pipeline consists of sentence segmentation, tool-based verification, and reflective reasoning with external resources, such as web search, calculators, and code interpreters. It was fine-tuned on synthetic detection trajectories. HaluAgent's performance was comparable to GPT-4’s on several benchmarks and also maintains strong generalization across domains in hallucination detection across different tasks, including open-domain QA, summarization, and dialogue generation.


\subsection{Self-Consistency-based Detection}
\begin{figure*}[t]
    \centering
    \includegraphics[width=\linewidth]{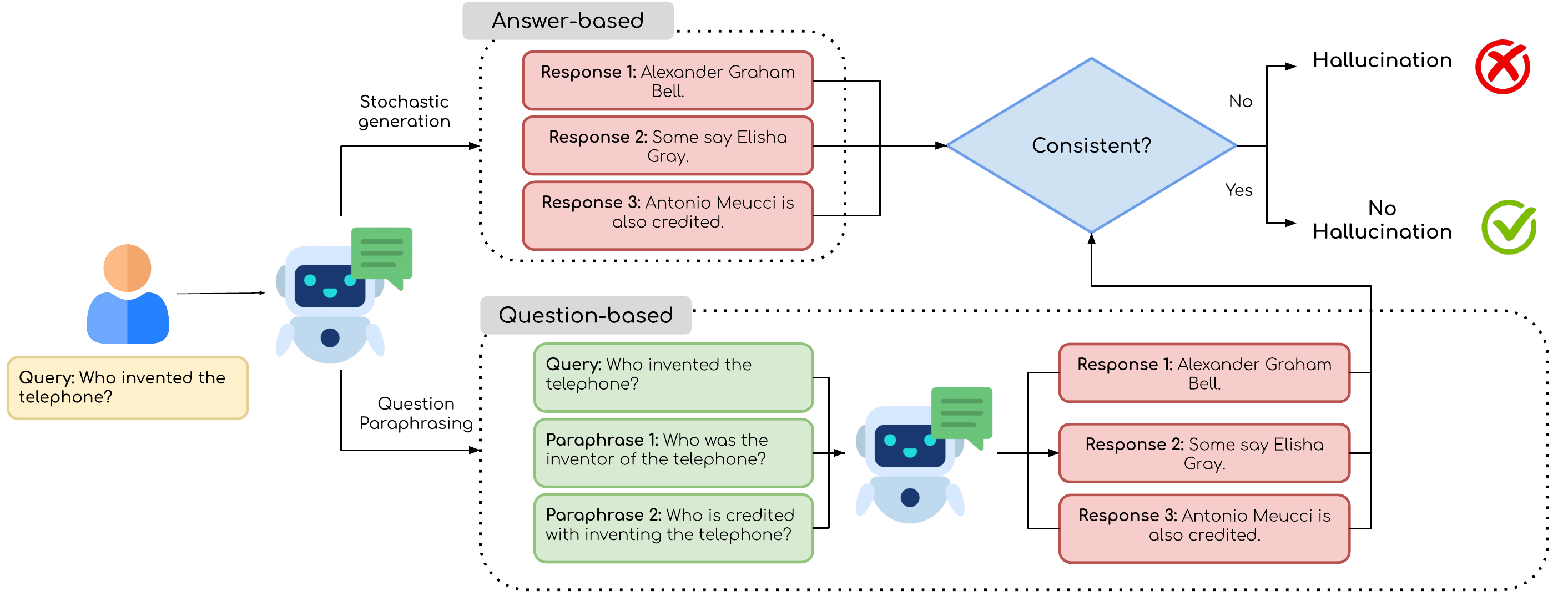}
    \caption{Examples of answer-based and question-based self-consistency methods for hallucination detection.}
    \label{fig:consistency}
\end{figure*}

Self-consistency is an unsupervised technique that improves LLMs' reasoning by generating multiple responses to a single prompt and assessing their internal consistency. Self-consistency does not require ground truth references, making it useful in open-ended and low-resource settings. Self-consistency-based techniques can be categorized into answer and question-based methods. Answer-based methods generate multiple answers to the same query using varied decoding hyperparameters, such as temperature and top-k. It then evaluates the model’s consistency across these responses. Question-based methods, on the other hand, generate answers to paraphrased versions of the same question and assess the model’s consistency across these different phrasings. Figure \ref{fig:consistency} illustrates the difference between answer and question-based self-consistency hallucination detection.

\textbf{Answer-based Methods.}
Manakul et al. \cite{manakul2023selfcheckgpt} introduced SelfCheckGPT, which utilizes self-consistency to detect LLMs' hallucination. This approach examines various techniques, including BERTScore similarity, question-answering consistency, n-gram probabilities, NLI, and prompt-based evaluation to assess the consistency of the generated text. The key idea of this approach is that if a response is factual, repeated queries to the same prompt with slight randomness should yield consistent responses, whereas hallucinated content would yield highly variable responses. Li et al. \cite{li2024think} introduced a comprehensive answer evaluation framework called think twice before trusting (T3). This approach prompts the model to generate multiple candidate answers, reflect on each one, and provide justifications. It then compares these answers to assess their relative trustworthiness, aiming to reduce overconfidence in incorrect outputs and improve the reliability of the final prediction. Experiments across multiple tasks and LLMs show that T3 substantially improves the AUROC for hallucination detection and calibration compared to prior self-consistency and confidence-based baselines. \textcolor{black}{More recently, SelfElicit \cite{liu2025long} detected hallucination in long-form text by decomposing them into statements and evaluates each one using calibration-based fact-checking. The model predicts True/False/Not Sure and the confidence of “False” serves as a hallucination score. The approach then elicits reflective thoughts conditioned on this evaluation and stores them in a knowledge hypergraph, which provides contextual semantic information for subsequent statements. A self-consistent, NLI-based conflict-resolution mechanism is applied to detect and mitigate contradictions, preventing hallucination snowballing.}

\textbf{Question-based Methods.}
SAC3 \cite{zhang2023sac3} extends self-consistency by incorporating semantic-aware cross-checking, including question perturbation (question rephrasing) and cross-model verification (output comparison across different LLMs) to detect points where self-consistency fails. These techniques address question-level hallucinations, where a model consistently generates incorrect but plausible responses, and model-level hallucinations, where there are inconsistencies between models. Yang et al. \cite{yang2025hallucination} further advance self-consistency-based detection with MetaQA. This approach utilizes metamorphic relations, such as synonym and antonym prompt mutations, to systematically alter queries and evaluate LLM responses for factual consistency. This technique outperformed SelfCheckGPT in zero-resource settings. Similarly, Xue et al. \cite{xue2025verify} present a two-stage methodology that integrates self-consistency with cross-model consistency. Their approach initially uses conventional self-consistency detection. Followed by a verifier LLM to validate ambiguous responses. 


\subsection{Detection Challenges}
Most hallucination detection approaches proposed in the literature depend on either uncertainty estimation or self-consistency verification to detect hallucination. 
The main hypothesis of uncertainty estimation is that a high uncertainty indicates that the model is guessing instead of relying on learned patterns. Therefore, the uncertainty-based detection approaches flag LLMs’ outputs with low confidence as potential hallucinations. 
However, multiple uncertainty-based hallucination detection approaches, especially those relying on token-level probability or entropy, tend to over-predict hallucination. Most of these approaches incorrectly flag too many outputs as hallucinations \cite{yeh2025can}, which leads to low precision, especially when hallucinations are sparse or when dealing with informative content, such as named entities. Moreover, these approaches fail in capturing hallucinated responses generated by models with high certainty \cite{simhi2025trust}. Other techniques, such as BTPROP \cite{hou2024probabilistic}, are extensive in terms of computational costs due to the recursive decomposition and multiple model queries.  


Self-consistency verification is an unsupervised technique that enhances the reasoning capabilities of LLMs by generating multiple responses to a single prompt and evaluating their internal consistency. Self-consistency-based hallucination detection approaches do not require ground truth references, making them useful in open-ended and low-resource settings.  However, the effectiveness of self-consistency hallucination detection depends on the diversity of prompts and sampling strategies. These methods may fail to detect inconsistencies that indicate hallucinations \cite{sriramanan2024llm} if the sampled responses lack diversity. Similarly, if an LLM is overconfident about a fact, self-consistency methods will fail to detect hallucination. In these cases, all sampled responses may agree on the same incorrect information, leading to high consistency scores for hallucinated content. Additionally, self-consistency approaches do not verify outputs against external knowledge or ground truth. As a result, they cannot catch hallucinations that are confidently and consistently produced by the model but are factually incorrect in the real world \cite{zhang2023sac3}.

In contrast to uncertainty and self-consistency-based hallucination detection approaches that do not require ground truth references, retrieval-based detection approaches depend on external knowledge retrieval to verify the generated LLM's outputs. 
These approaches aim to enhance factual consistency and mitigate the risk of unsupported content by grounding model outputs in verifiable information.
However, the effectiveness of retrieval-based detection is fundamentally limited by the quality of the retrieved documents \cite{zhang2025hallucination}. If the retrieved documents contain incomplete, outdated, or irrelevant information, the detector may either miss hallucinations or incorrectly flag accurate statements. Additionally, LLMs often struggle to select between parametric knowledge and retrieved sources, which can result in hallucinations \cite{genesis2025integrating}. Furthermore, incorporating real-time retrieval increases computational cost and response latency, which can pose challenges for interactive and real-time applications \cite{zhang2025hallucination}. 


Embedding-based hallucination detection techniques depend on measuring the semantic similarity between input, output, and external references. The main assumption of these embedding-based approaches is that a faithful model output should be semantically close to its sources in a shared embedding space.  
Although embedding-based approaches have demonstrated good hallucination detection performance, their reliance on internal model states imposes several limitations. Performance can be degraded on out-of-domain data, rare linguistic phenomena, or low-resource languages, where embedding models are less robust and attention structures are less informative \cite{tang2024we}. Moreover, the choice of which model layers or heads to extract embeddings from can significantly affect the detection performance. Although Oblovatny et al. \cite{oblovatny2025attention} introduced a robust method for fine-grained selection of attention heads, it does not fully solve the challenge of detecting subtle hallucinations. Another inherent limitation of embedding-based detection techniques is their inability to capture hallucinations stemming from external knowledge misalignments with real-world facts, especially if the model's training data are outdated or incomplete. 

Learning-based hallucination detection approaches leverage supervised or unsupervised models trained on annotated data to classify LLM outputs as either hallucinated or factual. By learning from data, these methods can generalize to a wide range of hallucination types across different domains and tasks. However, supervised learning algorithms require an extensive amount of labeled or synthetic data for training. Models trained on specific tasks or synthetic hallucination data may not generalize well to unseen hallucination types or out-of-domain tasks. Although unsupervised learning hallucination detection techniques offer advantages in terms of annotation-free operation and potential real-time application, they may struggle to generalize across different domains and hallucination cases \cite{chuang2024lookback}.

\textcolor{black}{The observed limitations in each category indicate that no single detection paradigm is sufficient in isolation. Hybrid strategies that combine internal confidence signals, consistency checks, and external evidence grounding are increasingly explored to balance scalability, factual reliability, and robustness across tasks and domains. However, hybrid techniques must also be selected carefully, as combining methods with similar failure modes may not necessarily improve robustness. For example, both uncertainty-based and self-consistency approaches struggle when models produce high-confidence but incorrect outputs. This failure mode is also observed in embedding- and attention-based similarity methods, which rely on internal representations rather than external grounding. The usefulness of the detection technique is highly task- and domain-dependent. For instance, attention- or embedding-based similarity signals tend to be more informative in tasks such as summarization and context-based QA, where hallucinations often manifest as faithfulness errors relative to a given source document. In contrast, for open-domain QA, hallucinations often arise from incorrect world knowledge rather than source misalignment, making internal similarity or attention patterns less reliable indicators of factual correctness. In such cases, retrieval-based or knowledge-grounded verification becomes more critical. These observations suggest that effective hallucination detection requires complementary combinations of methods that address different failure modes, rather than redundant hybrids that amplify shared weaknesses.}

\newcolumntype{Y}{>{\raggedright\arraybackslash}X}
\newcolumntype{C}{>{\centering\arraybackslash}p{0.9cm}} 
\newcolumntype{M}[1]{>{\centering\arraybackslash}m{#1}} 

\begin{table*}[t!]
\centering
\tiny
\renewcommand{\arraystretch}{1.2}
\setlength{\tabcolsep}{4pt}

\caption{
Comparison of hallucination detection techniques. 
Detection categories include retrieval-based (R.), uncertainty-based (U.), embedding-based (E.), learning-based (L.), and self-consistency (S.). 
Results are reported in \textbf{AUROC} metric; * indicates \textbf{F1-score}, and $\dagger$ indicates \textbf{Accuracy}.
}
\label{tab:detection-comparison}

\begin{tabularx}{\linewidth}{
  p{0.020\linewidth}
  p{0.030\linewidth}
  X
  *{5}{>{\centering\arraybackslash}p{0.025\linewidth}}
  *{10}{>{\centering\arraybackslash}p{0.03\linewidth}}
}
\toprule
\textbf{Ref} & \textbf{Year} & \textbf{Task} &
\multicolumn{5}{c}{\textbf{Detection}} &
\multicolumn{10}{c}{\textbf{Result Per Dataset}} \\
\cmidrule(lr){4-8}
\cmidrule(lr){9-18}
& & & \textbf{R.} & \textbf{U.} & \textbf{E.} & \textbf{L.} & \textbf{S.} &
\rotatebox{90}{\textbf{WikiBio}} & \rotatebox{90}{\textbf{TruthfulQA}} & \rotatebox{90}{\textbf{TriviaQA}} & \rotatebox{90}{\textbf{SQuAD}} & \rotatebox{90}{\textbf{NQ}} &
\rotatebox{90}{\textbf{HaluEval-QA}} & \rotatebox{90}{\textbf{HaluEval-Sum}} & \rotatebox{90}{\textbf{RagTruth}} & \rotatebox{90}{\textbf{CNN/DM}} & \rotatebox{90}{\textbf{HotpotQA}} \\
\midrule

\cite{wang2023hallucination} & 2023 & Bio. gen. & \cmark &  &  & \cmark &  & 74.2 & & & & & & & & & \\
\cite{mishra2024fine} & 2024 & Info-seeking & \cmark &  &  & \cmark &  & & & & & & & & & & \\
 
\cite{ajmal2025evaluating} & 2025 & QA & \cmark &  &  & \cmark &  & & & & & & & & & & \\

\cite{xu2025jointcq} & 2025 & QA & \cmark &  &  & \cmark &  & & & & & & & & & & \\
\cite{zhang2024knowhalu} & 2024 & QA, Summ. & \cmark &  &  & \cmark &  & & & & & &72.3* &68.5* & & & 72.1* \\
 
\cite{paudel2025hallucinot} & 2025 & QA, Summ., D2T & \cmark &  &  & \cmark &  & 83.7* & & & & & & & 85.0* & & \\

\midrule
\cite{guerreiro2023looking} & 2023 & MT &  & \cmark &  & \cmark &  & & & & & & & & & & \\
\cite{zhang2023enhancing} & 2023 & Summ. &  & \cmark &  & \cmark &  & 77.7 & & & & & & & & & \\
\cite{hou2024probabilistic} & 2024 & Summ. &  & \cmark &  &  & \cmark & 90.4* & & & & & & & & & \\
\cite{chen2025enhancing} & 2025 & Summ. &  & \cmark &  &  &  & 78.3 &  & & & & & & & & \\
\cite{yang2023improving} & 2023 & QA &  & \cmark &  & \cmark &  & & & & & & & & & & \\

\cite{shelmanov2025head} & 2025 & QA &  & \cmark & \cmark & \cmark &  & & & & & & & & & & \\

\cite{niu2025robust} & 2025 & QA &  & \cmark &  & \cmark &  & & & 90.3 & 83.7 & 86.7 & & & & & \\

\cite{tong2025halunet} & 2025 & QA &  & \cmark & \cmark & \cmark &  & & & 81.1 & 92.2 & 81.1 & & & & & \\

\cite{binkowski2025hallucination} & 2025 & QA &  & \cmark & \cmark & \cmark &  & & 82.9 & 88.9 & 79.5 & 82.7 & 87.4 & & & & \\
\cite{chuang2024lookback} & 2024 & QA, Summ. &  &  & \cmark & \cmark &  & & & & & & & & & & \\
\cite{farquhar2024detecting} & 2024 & QA, Summ. &  & \cmark & \cmark &  &  & & & & & & & & & & \\

\cite{samaga2026halluzig} & 2026 & QA, Summ. &  & \cmark &  &  &  & 73.3 & & & 73.0 & & & & & & \\
\cite{bazarova2025hallucination} & 2025 & QA, Summ. &  & \cmark &  &  &  & & & & 96.0 & & & & & 60.1 & \\
\cite{dasgupta2025hallushift} & 2025 & QA,Summ., Dlg. &  & \cmark & \cmark &  & & & 93.0 & 99.2 &  & 95.0 & 53.0 & & & & \\

\midrule
\cite{dale2023detecting} & 2023 & MT &  & \cmark & \cmark &  &  & & & & & & & & & & \\
\cite{nonkes2024leveraging} & 2024 & QA, Summ. &  & \cmark & \cmark &  &  & 63.0 & & & & & & & & & \\
\cite{hu2024embedding} & 2024 & QA, Dlg., Summ. &  &  & \cmark & \cmark &  & 89.8* & & & & 97.2$\dagger$ & 95.1$\dagger$ & & & & \\

\midrule
\cite{zhang2024prompt} & 2024 & QA &  &  & \cmark & \cmark &  & & & & & & & & & & \\

\cite{kong2025halugnn} & 2025 & QA &  & \cmark & \cmark & \cmark &  & & 68.8 & 93.0 & & & & & & & \\

\cite{park2025steer} & 2025 & QA &  &  & \cmark & \cmark & & & 88.7 & 87.2 & & 78.0  & & & & & \\
\cite{choi2023kcts} & 2023 & Summ., Dlg. & \cmark & \cmark &  & \cmark &  & & & & & & & & & & \\

\cite{su2024unsupervised} & 2024 & Text gen. &  &  & \cmark & \cmark &  & & & & & & & & & & \\
\cite{cheng2024small} & 2024 & QA, Text, Code & \cmark &  &  & \cmark & \cmark & & & & & & 83.8* & & & & \\ 

\cite{yamada2025light} & 2025 & QA, D2T, Summ. &  &  & \cmark & \cmark &  & & & & & & 60.4* & 63.6* & & & \\

\midrule
\cite{manakul2023selfcheckgpt} & 2023 & Bio. gen. &  &  &  &  & \cmark & 80.3 & & & & & & & & & \\
\cite{li2024think} & 2024 & QA &  &  & \cmark &  & \cmark & & & & & & & & & & \\
\cite{zhang2023sac3} & 2023 & QA &  &  &  &  & \cmark & & & & & 77.2 & & & & & 88.0 \\

\cite{yang2025hallucination} & 2025 & QA &  &  &  &  & \cmark & & & & & & & & & & 71.7 \\
\cite{liu2025long} & 2025 & QA &  &  &  & \cmark & \cmark & & & & & & & & & & \\
\cite{xue2025verify} & 2025 & QA, Bio. gen. &  &  &  & \cmark & \cmark & & & & & & & & & & \\

\bottomrule
\end{tabularx}
\end{table*}

\textcolor{black}{\section{Hallucination Explainability}}
\textcolor{black}{Hallucination explainability extends hallucination detection by explaining why they occurred and where they came from. It focuses on generating human-understandable reasons for a model's hallucinations \cite{zhao2024explainability}. In this survey, we divide the existing studies into model-internal explainability, which interprets hallucinations through the model’s internal behavior, and evidence-based explainability, which grounds explanations in evidence alignment and reasoning structure.}

\textcolor{black}{\subsection{Model-Derived Explainability}
Model-internal explainability treats hallucinations as a consequence of model-internal dynamics. These methods explain hallucination using signals extracted from model outputs or model-inferred quantities. Uncertainty is currently treated as a component of explainability in LLMs \cite{salvi2025explainability}. Huang et al. \cite{huang2025reppl} introduced RePPL, which attributes hallucinations to token-level uncertainty derived from instability in semantic propagation through attention layers and low-confidence generation decisions. In another direction, the hallucination explanation is learned during hallucination detection. HuDEx \cite{lee2025hudex} is trained to justify its hallucination judgments directly using natural language, where it provides reasoning behind these judgments. Similarly, Xie et al. \cite{xie2025improving} introduced FENCE, which is trained to identify factual errors at the claim level. The model classifies claims as supported, contradictory, or unverified and generates explanatory critiques. Retrieval is used in a supporting role, providing external evidence from search engines, knowledge bases, and knowledge graphs to inform the evaluator’s judgments.}

\textcolor{black}{\subsection{Grounding Explainability}
Grounding explainability refers to methods that explain hallucinations by examining how generated content aligns or fails to align with external evidence. Grounding explainability is usually based on source documents, factual references, or claim decomposition. They identify hallucinations by revealing unsupported statements, contradictions, missing evidence, or broken inference chains, often through evidence highlighting, entailment checking, or structured reasoning verification. One prominent direction for grounding-based hallucination explainability is through a relational mapping of evidence to text. HaluCheck \cite{heo2025halucheck} provided explainability through evidence-grounded, sentence-level verification. The proposed technique decomposes model outputs into atomic facts. Then, it retrieves supporting documents from external knowledge sources, and uses NLI models to determine whether each fact is entailed. Hallucinations are highlighted directly, which allows users to see which specific statements lack evidence. Likewise, Chen et al. \cite{chen2025explainable} introduced HaluMap, which constructs a segment-level matrix of entailment and contradiction relations between source inputs and generated text using NLI models. HaluMap produces a heatmap-like visualization that highlights which parts of the output are unsupported or contradicted by the source. To improve robustness, they further propose SelfHaluMap, which is a calibration mechanism that estimates background inconsistency within the source document itself and reduces noise in the final explanation. In another direction, Hu et al. \cite{hu2024slm} introduced a two-stage framework that combines a small language model for rapid hallucination detection with an LLM-based constrained reasoner that generates detailed explanations for detected cases. Their approach dramatically reduces inconsistencies between detection and explanation, achieving a high F1 score for identifying inconsistent rationales. In another study, Orshansky et al. \cite{orshansky2025hallutree} focused on reasoning-structured explanations. They proposed HalluTree, which decomposes summaries into subclaims and organizes verification results into a hierarchical claim tree. Subclaims are categorized as extractive or inferential. Extractive subclaims are directly verifiable against the source, whereas inferential subclaims requires multi-hop reasoning. Extractive claims are checked with lightweight NLI models. On the other hand, inferential claims initiate a reasoning path, where an LLM proposes a set of supporting facts from the source text, logical or mathematical reasoning, or general knowledge. Later, they are organized into a coherent reasoning chain. An LLM then evaluates the claim’s groundedness using CoT reasoning. Both the supporting facts and the reasoning trace are attached into a verification tree to provide an explicit, explainable rationale. Similarly, Galitsky and Rybalov \cite{galitsky2025information} leveraged information gain with abductive reasoning to explain reasoning-based hallucinations. The proposed technique measures how much a model’s claim shifts the probability distribution away from what is supported by the source. It then checks whether any minimal, plausible hypothesis could logically justify that shift. If no simple abductive explanation exists, or only overly complex ones do, the claim is labeled a hallucination, and the missing or faulty reasoning path is explicitly identified.}

\textcolor{black}{\subsection{Explainability Challenges}
Despite recent progress in developing explainability methods for hallucinations, there remain significant open challenges that limit their effectiveness. Unlike classification labels, there is no universally agreed-upon correct explanation for why a model hallucinates in a given instance. Explanations often rely on proxy signals such as attention patterns, structural irregularities, or alignments with external evidence, which may not fully capture the underlying model's reasoning. This makes it difficult to evaluate the quality, faithfulness, and completeness of explanations produced by explainability methods. Moreover, in techniques that rely on reasoning trees and topological features, there is a gap between producing explanations that are accurate and faithful to the model’s internal mechanisms and those that are interpretable to humans, especially non-experts. This is because such techniques can be too technical for end users, while simpler explanations may omit key reasoning details, which reduces transparency. Those techniques are also computationally expensive, which limits their practicality for long texts, real-time applications, or large corpora. Furthermore, there is a lack of standardized benchmarks and metrics for evaluating explanation quality. Without agreed standards, comparing different methods and measuring improvement is difficult, which slows progress.}

\section{Hallucination Mitigation}
\label{sec_mitigation}
\begin{figure*}[ht]
    \centering
    \includegraphics[width=0.8\linewidth]{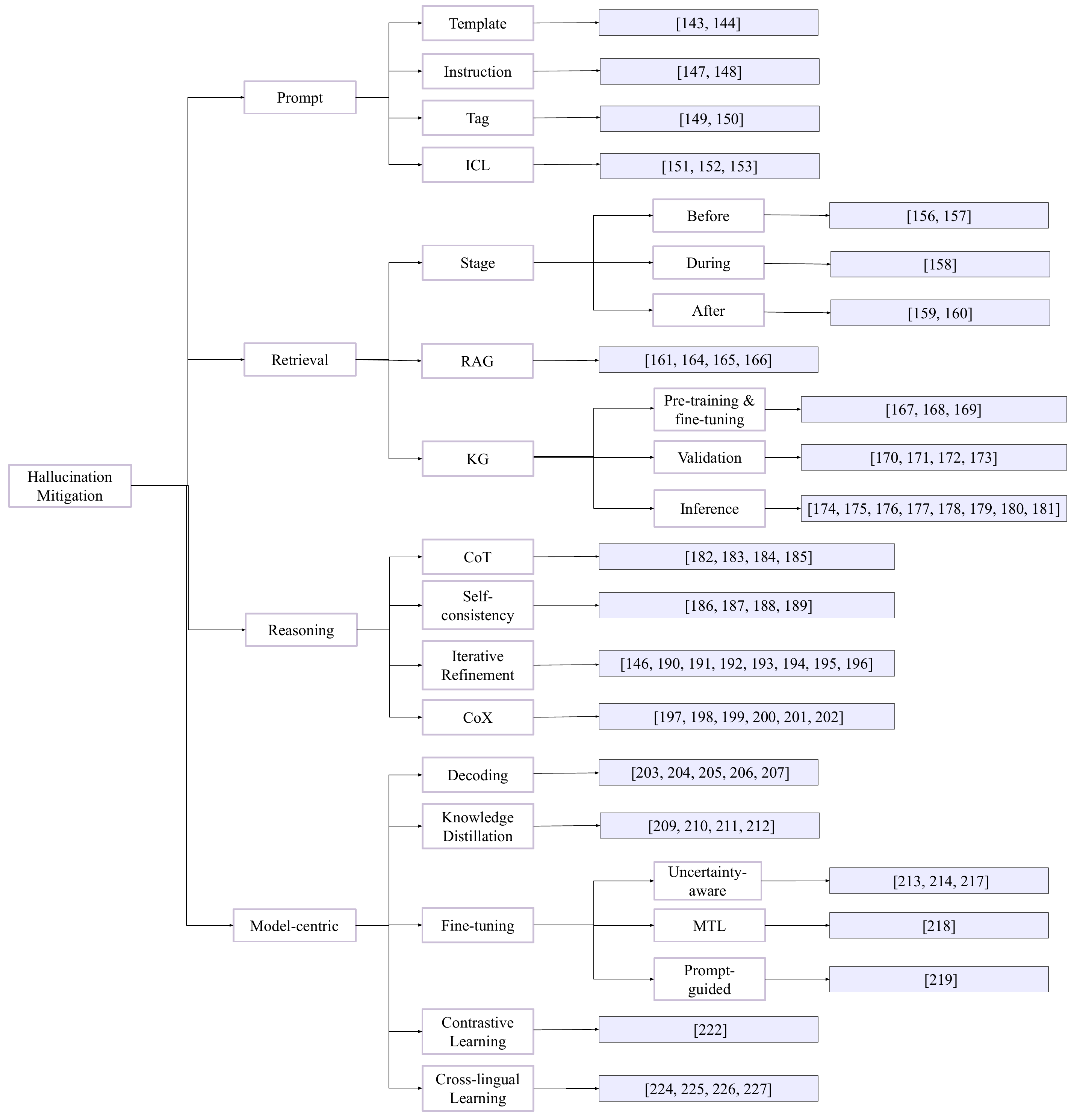}
    \caption{The taxonomy of hallucination mitigation methods.}
    \label{fig:mitigation}
\end{figure*}

Hallucination mitigation aims to reduce or prevent the emergence of factually inaccurate, ungrounded, and contextually inconsistent responses in LLMs' outputs. In contrast to hallucination detection, which focuses on detecting hallucinatory outputs, mitigation seeks to make modifications to LLMs to deliver accurate responses. Mitigating hallucinations is a crucial priority in the advancement and implementation of LLM to ensure LLM output remains factual, trustworthy, and safe. As shown in Figure \ref{fig:mitigation}, this survey categorizes the hallucination mitigation techniques proposed in the literature into prompt-based, retrieval-based, reasoning-based, and model-centric training and adaptation-based techniques. Table \ref{tab:mitigation-comparison} summarizes the mitigation studies with the datasets used in each study and the metrics employed to evaluate their techniques. \textcolor{black}{Moreover, Table \ref{tab:qa_sota} presents results on the four most widely used datasets for hallucination mitigation.}

\textcolor{black}{Hallucination mitigation is followed by post‑repair methods, which repair the detected hallucination spans with minimal alteration to the rest of the output. Targeted correction methods operate at the span or entity level. Once specific tokens, entities, or propositions are flagged as hallucinated, the model regenerates or edits only those parts, often under additional constraints or with external evidence \cite{vladika2025correcting,zhang2024truthx}. On the other hand, partial regeneration methods relax the granularity from spans to larger segments such as sentences, paragraphs, or specific reasoning steps \cite{vladika2025correcting,kamoi2024can}. The system selectively regenerates only the hallucination segments, while preserving the rest of the answer.}

\subsection{Prompt-Based Techniques}
Prompt engineering has emerged as an effective way to guide and regulate LLM behavior. A well-constructed prompt can play an essential role in mitigating hallucinations \cite{sahoo2024systematic}. The notion of prompt design originated with the advent of GPT-2 \cite{radford2019rewon} and GPT-3 \cite{brown2020language}, when researchers began exploring the capability of pre-trained LLMs to perform various tasks solely by altering the input prompt, without the need for task-specific fine-tuning. In his section, we classify prompt design strategies for hallucination mitigation into four major categories: template, instruction, tag, and in-context learning (ICL)-based prompts.

\textbf{Template-based prompts.} Employ predefined structures with placeholders, which allow for standardized input-output formats across different examples. 
Recent research supports the effectiveness of template-based prompts in mitigating hallucinations of LLMs \cite{jiang2023ai,yang2023refgpt}. Jiang et al. \cite{jiang2023ai} introduced a structured prompt-template framework for AI-generated news articles, where key components, such as introduction, event, and argument, are used to guide the model in producing factually accurate content. The proposed framework also involves a post-checking process that verifies compliance between the structured input and output, thereby reducing the occurrence of hallucinated information. Similarly, RefGPT \cite{yang2023refgpt} introduced structured prompting for dialogue generation using three components: reference selection, basic prompting, and dialogue settings. This setup ensures that the model draws solely from specified references, which minimizes reliance on parametric knowledge.

\textbf{Instruction-based prompts.} Provide explicit and clear instructions to an LLM to guide its output towards the intended result using natural language. 
Studies \cite{wei2022chain, madaan2023self} have shown that instruction-based prompting, when coupled with techniques such as CoT reasoning or iterative refinement, significantly reduces hallucinations by constraining model responses to align with facts and rationality. Moreover, LLM fine-tuning on instruction-based datasets enhances its ability to follow complex instructions and reduce errors \cite{ouyang2022training}. Kim et al. \cite{kim2024self} proposed SELF-EXPERTISE, a method that generates knowledge-based instruction datasets rather than relying on LLMs' parametric knowledge. It extracts factual knowledge from seed dataset outputs to create structured instruction, input, and output pairs, ensuring factual accuracy. Additionally, system instructions provide explicit guidance for generating logically sound responses, particularly in specialized domains, such as law.

\textbf{Tag-based prompts.} These approaches depend on tailoring LLMs to perform certain tasks by integrating additional information "tags" into the input. These tags or markers are incorporated into the prompt to direct the model's focus and response formulation. 
Feldman et al. \cite{feldman2023trapping} conducted a study on the impact of tagged prompting on LLM hallucination mitigation. Their research demonstrated a remarkable success rate of 98.88\% in eliminating fabricated information when using context-embedded tags. Similarly, Penkov \cite{penkov2024mitigating} proposed a method that integrates domain-specific tools, such as BioBERT and ChEBI, with tagged prompting. Their approach focused on anchoring LLM outputs to verifiable facts, particularly in specialized fields, such as biomedicine. The study showed that this combination of techniques could substantially reduce hallucinations by providing a semantic framework for the model to adhere to during response generation.


\textbf{ICL prompts.} Refer to an alternative learning paradigm in which the prompt contains some examples for the LLM to follow to complete the task given. When combined with other techniques, ICL represents a valuable approach to mitigating hallucinations in LLMs. By leveraging the model's ability to adapt to new tasks and information without parameter updates, ICL can potentially improve factual accuracy and reduce erroneous outputs \cite{liang2024thames}. Moreover, Zhang et al. \cite{zhang2023summit} showed that ICL with iterative refinement enhances summary faithfulness, demonstrating its advantages in faithfulness, controllability, and overall quality. Similarly, Vu et al. \cite{vu2023freshllms} showed that few-shot prompts containing up-to-date information can improve factual accuracy.

\subsection{Retrieval-Based Techniques}
Retrieving external knowledge for LLMs can enhance their factual grounding, contextual relevance, and overall accuracy. The information contained in the model parameters during pre-training is known as parametric knowledge. This knowledge is static and may not encompass the most recent or domain-specific information. To overcome this limitation, non-parametric knowledge can be incorporated through retrieval-based techniques. In the context of LLMs, knowledge retrieval enables them to dynamically access external, verified sources of data, such as search engines, databases, and specific text corpora. By interacting with these reliable sources, models can consult up-to-date and domain-specific data during generation, which in turn improves the factual accuracy of their outputs \cite{guu2020retrieval,lee2019latent}. Retrieval-based techniques can be classified into three categories based on the phase of accessing external resources: before, during, and after generation \cite{huang2023survey}. 
We used these knowledge retrieval stages to categorize retrieval-based techniques proposed for hallucination mitigation.


\textbf{Retrieving knowledge before generation} involves querying external resources to find relevant information related to the user's query before generating the response. This approach helps mitigate hallucination by constraining the model to generate responses based on verified sources rather than relying solely on its internal knowledge. This ensures that the generated content is grounded in factual data. Ni et al. \cite{ni2023chatreport} developed a framework for summarizing sustainability reports aligned with the Task Force on Climate-related Financial Disclosures guidelines. Hallucinations are mitigated by restricting the generated summarization to use only the retrieved reports. The system uses carefully designed queries to retrieve relevant content from the reports and prompts the LLM to summarize it in alignment with the guidelines. Peng et al. \cite{peng2023check} introduced LLM-AUGMENTER that mitigates hallucination in GPT-3.5 for information-seeking dialogue and open-domain QA. It incorporates external knowledge and refines prompts before generation. The LLM-AUGMENTER optimizes the policy module using RL to retrieve and consolidate external knowledge, refine prompts, and iteratively improve the quality of the answers based on utility input. 

\textbf{Knowledge retrieval during generation} involves querying external knowledge during the process where an LLM is generating the response. The retrieved information is combined with the original query to provide context for the LLM. Nathani et al. \cite{nathani2023maf} proposed a technique that iteratively refines hallucinated text and reasoning errors using multiple feedback sources. The framework begins with a base model that generates an initial response. This response then undergoes an iterative refinement process, where feedback from various modules is applied across multiple iterations. The feedback modules are designed to address specific error categories, including factual inaccuracies, commonsense mistakes, and redundancy. Feedback can be provided by pre-trained LLMs or external tools. Then, a refiner model, which is the same as the base model, utilizes the feedback to refine its response.

\textbf{Knowledge retrieval after generation} involves engaging in a fact-checking or validation process by cross-referencing the initial response against a knowledge source. Gao et al. \cite{gao2023rarr} aimed to detect and mitigate hallucinations in LLMs after generation by finding attribution and post-editing unsupported content. The authors introduced RARR, a system that extracts evidence from external sources to validate the output text's dependability. The RARR framework consists of two main phases: research and revision. In the research phase, a query generator formulates inquiries regarding various aspects of the text, while a retriever seeks supporting evidence. In the revision phase, an agreement model identifies discrepancies between the text and the evidence, followed by an edit model that revises the text if necessary. The process concludes with an attribution report linking the revised content to its sources. Similarly, Huo et al. \cite{huo2023retrieving} proposed a validation method that enables an LLM to cross-check its generated answers against retrieved evidence. The authors suggested a retrieval pipeline consisting of sparse retrieval, dense retrieval, and neural re-rankers. This pipeline has been employed in two ways. First, the generated response and query were used to retrieve external evidence, which was then employed to validate the output. In the second way, the LLM extracted multiple factual statements from the response, each of which was independently verified against retrieved evidence.

Recent approaches \cite{lewis2020retrieval, song2024rag, hogan2021knowledge, agrawal2024can} that depend on retrieving knowledge after generation have significantly integrated RAG and KG for hallucination mitigation in LLMs. RAG and KG represent effective solutions in helping LLMs avoid hallucinations by grounding their responses on reliable and fact-checked information. RAG enhances the output of an LLM with dynamic retrieval and incorporation of external and up-to-date knowledge, which decreases dependency on static, parametric knowledge and improves adaptability in real-world applications. KGs, on the other hand, offer structured, relationship-centric representations of facts that capture entities and their relations. They can be integrated at various stages from LLMs' pre-training and fine-tuning to inference and validation. This technique enables models to navigate complex relationships and facts accurately. Both techniques complement each other to improve LLM capabilities and contribute to more reliable factual consistency outputs, which enables LLMs to generate text with better accuracy in various tasks.

\subsubsection{Retrieval Augmented Generation}
RAG enhances LLMs by retrieving relevant external documents at inference time, which enables the LLM to ground its responses in factual evidence beyond its parametric training data \cite{lewis2020retrieval}. It consists of two components, a retriever and a generator, which work together in a retrieve-then-read pipeline. Unlike traditional retrieval techniques, RAG utilizes pre-trained components that are already loaded with extensive knowledge. Consequently, it can immediately access and integrate a broad range of data without the need for additional training.

RAG-HAT \cite{song2024rag}, a hallucination-aware tuning pipeline, employs a three-step process: detection, rewriting, and mitigation. It uses a fine-tuned detection model to identify hallucinations and provide detailed descriptions, which are then used to guide GPT-4 in revising the RAG output. The revised outputs are used to create a preference dataset for direct preference optimization training, resulting in reduced hallucination rates and improved answer quality. TRAQ \cite{li2024traq} combines RAG with conformal prediction to provide statistical correctness guarantees for QA. It applies conformal prediction to both the retrieval and LMs, which generates sets of passages that contain relevant information and answers to generate the correct response. Furthermore, Ayala et al. \cite{ayala2024reducing} demonstrated that using a well-trained retriever can significantly decrease hallucination rates, particularly in out-of-domain settings. This approach allows for the deployment of smaller LLMs without compromising performance, which makes it especially useful for enterprise applications with resource constraints. In another study, Rowen et al. \cite{ding2024retrieve} proposed a framework that enhances LLMs with an adaptive retrieval augmentation process tailored to address hallucinated outputs. The authors employed a consistency-based hallucination detection module, which assesses the model's uncertainty regarding the input query by evaluating the semantic inconsistencies in various responses generated across different languages or models. When high uncertainties in the responses are detected, it activates the retrieval of external information to rectify the model outputs.

\subsubsection{Knowledge Graph}
KG, also known as semantic networks, systematically arranges information in a structured manner to establish relationships among real-world entities \cite{hogan2021knowledge}. It serves as a powerful tool for storing and querying complex information in a way that mirrors how humans conceptualize knowledge. KG has been used by researchers to mitigate hallucination by incorporating it with pre-training and fine-tuning, validation, and inference stages of the LLM development cycle \cite{agrawal2024can}.


\textbf{Pre-training and fine-tuning stages.} KG helps mitigate hallucinations introduced during pre-training and fine-tuning stages by providing LLMs with access to accurate, structured, and up-to-date factual information, thereby grounding generation in verifiable knowledge. It provides structured data about entities and their relationships, which enhances LLMs’ comprehension and helps them generate language that accurately reflects real-world complexities. ERNIE 3.0 \cite{sun2021ernie} integrates KGs by linking words to entities through knowledge masking, allowing the model to capture both contextual and relational information more effectively. Similarly, KGLM \cite{youn2023kglm} leverages KG triples in pre-training and fine-tuning stages using an additional entity-relation-type embedding layer. This layer strengthens the model’s ability to understand KG structures, recall factual data, and improve accuracy in knowledge-based tasks. KG-Adapter \cite{tian2024kg} integrates KGs through parameter-efficient fine-tuning using dual-perspective adapter modules. This approach encodes node-centered and relation-centered KG structures while reducing knowledge conflicts by 38\% compared to prompt-based methods, which require only 28M trained parameters for 7B-scale LLMs.


\textbf{Validation stage.} KGs can add a fact-checking layer, which provides a comprehensive explanation to justify the LLM's decision. FOLK \cite{wang2023explainable} is a fact-verification method that employs KGs using first-order-logic predicates to verify claims in online misinformation. It can also provide explanations of its findings to help human fact-checkers comprehend and evaluate the model's outputs. Emerging directions include neuro-symbolic integration \cite{ji2024retrieval}, which combines neural retrieval with logical KG reasoning, and autonomous retrofitting \cite{regino2025can}, which iteratively aligns LLM outputs with KG-derived constraints. An approach for generating faithful text from KGs with noisy reference text has been introduced in \cite{hashem2023generating}. This method incorporates contrastive learning to enhance the model's ability to differentiate between faithful and hallucinated information. It encourages the decoder to generate text that aligns with the input graph. Additionally, it employs a controllable text generation technique that allows the decoder to manage the level of hallucination in the generated text.

\textbf{Inference stage.} 
KGs can also mitigate hallucinations in the inference stage by employing KG-augmented retrieval. This allows LLMs to retrieve the appropriate subgraph for answering a query grounded by real-world facts. KAPING \cite{baek2023knowledge} retrieves related triples from KGs for zero-shot QA by matching entities in questions. Similarly, StructGPT \cite{jiang2023structgpt} enhances LLM responses using data from KGs, tables, and databases and employing structured queries for information retrieval. \textcolor{black}{In another study, KG is used to enhance evidence acquisition \cite{yang2025curiousllm}. CuriousLLM \cite{yang2025curiousllm} is proposed, which builds and traverses a passage-level KG where nodes represent documents and edges reflect learned semantic connections. The LLM agent generates follow-up questions that guide traversal over this graph to locate missing evidence across documents. The proposed technique actively identifies what information is still missing and continues retrieval until sufficient support is found, or terminates early when evidence is complete.}

For complex reasoning tasks, IRCoT \cite{trivedi2023interleaving} integrates KGs with CoT to iteratively guide retrieval and reasoning for multi-step questions. Likewise, RoG \cite{luo2023reasoning} uses KGs to construct reliable reasoning pathways grounded in diverse relationships to improve accuracy. \textcolor{black}{Moreover, PoG \cite{tan2025paths} retrieves a question-specific subgraph for each query, explores multi-hop reasoning paths between relevant entities, and feeds these structured paths into the LLM as evidence for answer generation. Hallucinations are mitigated by constraining the model to reason over verifiable KG paths, pruning noisy or irrelevant branches, and verifying whether the retrieved paths provide sufficient support before producing an answer.} For real-time applications, FactGenius \cite{gautam2024factgenius} combines LLM-based connection filtering with Levenshtein distance validation, which improved fact verification F1-scores by 12\% on the FactKG benchmark through fuzzy relation mining. Moreover, KG-based retrofitting (KGR) \cite{guan2024mitigating} incorporates LLMs with KGs to mitigate factual hallucination during the reasoning process. Unlike previous methods that rely solely on user input to query the knowledge graph, KGR retrofits LLM initial draft responses using factual knowledge stored in KGs. The framework leverages LLMs to extract, select, validate, and retrofit factual statements in model-generated responses, enabling an autonomous knowledge-verification and refinement process without additional manual effort.


\subsection{Reasoning-Based Techniques}
Reasoning-based mitigation techniques in LLMs aim to address complex tasks that require logical and step-by-step thinking. Traditional LLMs often struggle with intricate reasoning and may produce factually inaccurate or misleading outputs when handling tasks that require multi-step thinking or inference. To mitigate this, several reasoning-based methods, such as CoT reasoning, self-consistency, iterative refinement, and Chain-of-X (CoX) prompting, were used to improve reasoning accuracy and reduce hallucinations.

\textbf{CoT Reasoning.} CoT is an improved version of few-shot prompting \cite{wei2022chain} that follows a step-by-step approach in natural language to decompose multi-step problems into intermediate steps. This method provides an interpretable view of the model's reasoning process, which enables better insight into how the model arrives at its conclusions. Zhao et al. \cite{zhao2024enhancing} proposed a zero-shot CoT reasoning method by incorporating logical thoughts (LoT), which employs symbolic logic to validate and revise each step. Therefore, it addresses and mitigates hallucinations arising from logical flaws directly. Sultan et al. \cite{sultan2024structured} introduced Structured CoT (SCoT), which uses a structured state-machine methodology to systematically oversee content reading, utterance generation, and hallucination mitigation phases. This method improves the model's faithfulness and reduces hallucinations in content-grounded tasks. Li et al. \cite{li2025structured} further adapted the SCoT approach specifically for code generation tasks, introducing programming structures, such as sequential, branching, and looping steps, as intermediate representations. Their approach explicitly guides LLMs to follow structured programming logic, which enhances the logical correctness, clarity, and readability of the generated code. This structured reasoning substantially mitigates hallucinations and improves code accuracy, outperforming traditional CoT prompting by a considerable margin across multiple programming benchmarks. \textcolor{black}{Li et al. \cite{li2025mitigating} improved the causal reasoning capabilities of LLMs by introducing CDCR-SFT. The authors fine-tune LLMs to explicitly construct causal graphs and reason over them before producing answers. By shifting reasoning from surface token patterns to structured causal relationships, the model reduces logically inconsistent hallucinations that arise from flawed reasoning chains.}

\textbf{Self-Consistency.} Building on the success of CoT, Wang et al. \cite{wang2022self} introduced a self-consistency approach, which replaces the decoding strategy of CoT. Unlike the original CoT method, the authors proposed the concept of generating multiple reasoning chains for a certain topic. The final answer is determined by conducting a majority vote or by selecting the most consistent response from several chains. Li et al. \cite{li2022making} extended self-consistency by proposing DIVERSE. This method generates multiple reasoning pathways through diverse prompts, systematically validates each reasoning step, and utilizes weighted voting according to step consistency. The step-aware approach significantly reduces hallucinations by isolating and rectifying errors at the step level rather than the whole chain. Similarly, Liang et al. \cite{liang2024learning} proposed an RL from knowledge feedback (RLKF) technique, which integrates self-consistency and internal knowledge state assessments. RLKF enhances the model's ability to resist factual hallucinations by reinforcing the consistency between the internal knowledge state and external outputs. Xu et al. \cite{xu2024sayself} proposed SaySelf, a method for training LLMs to produce self-reflective rationales conditioned on inconsistencies across multiple sampled reasoning trajectories. SaySelf explicitly prompts the model to provide precise confidence estimates and rationales that detail its knowledge gaps and uncertainty. This method substantially improves confidence calibration and effectively mitigates hallucinations.

\textbf{Iterative Refinement. }
CoT has substantially improved the reasoning abilities of LLMs, thus reducing hallucination. Nevertheless, if CoT initiates the sequence with erroneous reasoning, it fails to rectify the reasoning errors or factual inaccuracies that may occur during the reasoning process. This limitation often leads to hallucinations. Inspired by the way people edit their written content, Madaan et al. \cite{madaan2023self} proposed Self-Refine to enhance the generated output by an LLM through iterative refinement, which depends on domain-specific data, external supervision, and RL. However, these techniques require large annotated datasets that are unavailable for several domains. Self-Refine leverages the concept of LLMs as self-reviewers by using a single LLM that acts as a generator, refiner, and feedback provider, relying solely on prompt designs without the need for fine-tuning. Self-Refine depends on an appropriate LLM and three prompts (for initial response generation, feedback, and refinement). It works by generating an initial output based on an input sequence, providing feedback on the output, then subsequently refining the output in accordance with the self-generated feedback. It alternates between feedback and refining until a specified condition is achieved. Similarly, Shinn et al. \cite{shinn2024reflexion} proposed Reflexion method that involves finding the subsequent optimal solution in planning using ReAct \cite{yao2022react}. The Reflexion uses a long-lasting memory that allows an agent to recognize its own mistakes and autonomously derive insights from its mistakes and iteratively adapt its behavior over time. Instead of using RL, Reflexion leverages an external verbal feedback mechanism. This technique operates by employing an actor that generates an output and an evaluator to evaluate the generated output. The process continues until the evaluator thinks the agent’s output is correct or when a maximum number of trials is reached. In contrast to Self-Refine, Reflexion explicitly integrates external evaluations and memory mechanisms, which enhances learning effectiveness over both short-term and long-term experiences. Paul et al. \cite{paul2024refiner} developed REFINER by applying structured intermediate criticism from a critic model. The critic model critiques single reasoning steps and allows iterative refinement by identifying and correcting inaccuracies throughout the reasoning process. Likewise, Lee et al. \cite{lee2024ask} introduced the Ask, Assess, and Refine (A2R) methodology, which methodically assesses outputs for factual accuracy and hallucinations through metric-based feedback. A2R employs natural language feedback from metric assessments, which progressively enhances responses according to metrics like accuracy, fluency, and citation precision to reduce hallucination. In another study, Zhang et al. \cite{zhang2024prefer} introduced PREFER, a refinement-focused ensemble method, which works by continuously merging and refining prompts through identifying the limitations of previous iterations, enhancing model stability and overall output quality. In addition, Huang et al. \cite{huang2023large} demonstrated that LLMs can enhance their reasoning capabilities by generating high-confidence rationale-augmented responses. Their approach integrates unsupervised iterative refinement with self-consistency and feedback-based fine-tuning, which significantly improved reasoning performance. \textcolor{black}{Cheng et al. \cite{cheng2025think} proposed HaluSearch models generation as a slow, step-by-step reasoning search using tree search (e.g., MCTS). Each intermediate reasoning step is scored by a reward model that estimates hallucination risk, and the model dynamically switches between fast and slow thinking. This prevents early reasoning errors from propagating and promotes more reliable reasoning paths.}

\textbf{CoX Reasoning.} The sequential thought structure of CoT served as the inspiration for several techniques referred to as CoX techniques in this survey. They have been designed to tackle problems in various tasks and domains by merging with other techniques, such as iterative refinement and knowledge retrieval, to mitigate hallucination. These techniques include chain-of-verification (CoVE), chain-of-natural language inference (CoNLI), chain-of-question (CoQ), chain-of-knowledge (CoK), and chain-of-notes (CoN). 

The \textit{CoVE prompting} approach, inspired by the notable success of the reasoning chains and self-refine paradigms, aims to mitigate hallucination by enabling an LLM to formulate verifiable questions to authenticate its reasoning \cite{dhuliawala-etal-2024-chain}. The model initially produces a baseline response to serve as a reference for further improvements. The LLM then generates verification questions to check the factual accuracy of its statements. Several verification execution strategies have been proposed, such as joint, 2-step, factored, and factor+revise methods.  In the joint execution technique, the LLM generates verification questions and answers within a single prompt, while the 2-step method generates them in two separate prompts. Conversely, the factored technique handles each verification question in a separate prompt, while factor+revise assesses the coherence between the baseline response and the verification answers to produce the final verified response. The reported results show that the factor+revise method yields the strongest overall factuality score compared with other techniques.

 The \textit{CoNLI prompting} approach uses a hierarchical inference-based framework to detect and mitigate ungrounded hallucinations in LLMs \cite{lei2023chain}. A detection agent breaks down the baseline response into claims at the sentence and entity levels. Each sentence or entity is treated as a hypothesis, which is then verified using NLI against the original text. A mitigation agent subsequently corrects detected hallucinations by post-editing the baseline output to address inconsistencies. The results show that CoNLI-GPT-4 consistently outperforms other approaches across multiple datasets in both hallucination detection and mitigation.
The \textit{CoQ prompting} approach enhances CoT by breaking down complex questions into multiple sub-questions and incorporating a knowledge retrieval mechanism \cite{huang2024coq}. CoQ requires that each reasoning step be supported by at least one retrieved knowledge source. This approach prevents hallucination during reasoning and ensures the model's cognitive steps are grounded in factual information. On average, results show that CoQ reduces factual inaccuracies by 31\% compared to CoT alone and by 38\% relative to the two most prevalent LLMs.

 The \textit{CoK prompting} approach uses KGs to enable LLMs to perform knowledge reasoning \cite{zhang2024chain}. CoK employs two methodologies: data building and model learning. Data building involves three steps: rule mining, which derives rules from KG triples; knowledge selection, which selects appropriate knowledge for CoK data construction; and sample generation, which converts knowledge into natural language. Model learning uses both conventional behavior cloning and a trial-and-error approach to mitigate rule overfitting, which can cause hallucination. This trial-and-error mechanism enables the model to explore various reasoning paths and apply alternative rules when critical information is absent. In out-of-domain evaluation, CoK demonstrated superior performance in knowledge reasoning compared to baseline methods. Another CoK technique \cite{li2023chain} mitigates hallucination by dynamically incorporating knowledge from heterogeneous sources, including structured and unstructured data. This framework comprises three stages: reasoning preparation, dynamic knowledge adaptation, and answer unification. Initially, CoK formulates several rationales and selects answers lacking majority consensus for further processing using CoT with self-consistency. An adaptive query generator dynamically formulates queries tailored for various knowledge sources to refine the rationales, correcting each step to address error propagation. Experiments on knowledge-intensive tasks, including factual, medical, physics, and biological domains, show that this CoK framework substantially improves LLM performance.

 The \textit{CoN prompting} approach is an approach to improve the relevancy of retrieval-augmented LMs (RALMs) \cite{yu2023chain}. CoN addresses two primary challenges: handling noisy and irrelevant information and recognizing the absence of sufficient knowledge. This enables RALMs to assess their knowledge sufficiency and reply with "unknown" when information is lacking. The main innovation of CoN is the generation of brief reading notes for each document retrieved by the model. These notes summarize key points in each document, allowing the model to evaluate their relevance to the input query. When retrieved documents lack relevant details, CoN can instruct the model to acknowledge its limitations by answering "unknown" or providing justifications based on available data. Evaluations across open-domain QA datasets show significant improvements, with a 7.9\% average increase in exact match scores for noisy retrieved documents and a 10.5\% increase in rejection rates for out-of-scope questions.


\subsection{Model-centric Training and Adaptation}
These processes involve refining model architectures, adjusting training strategies, and integrating advanced techniques that make outputs coherent, contextually relevant, and fact-based. This survey categorizes the model-centric training and adaptation-based approaches for hallucination mitigation into four main strategies: optimizing decoding methods, leveraging knowledge distillation, applying supervised fine-tuning, and adopting self-learning techniques.

\textbf{Decoding Strategies.}
Decoding is the mechanism used to convert encoded representations of the LLMs' output into comprehensible text. During decoding, the model iteratively selects tokens from its vocabulary, which build contextually relevant and syntactically accurate sentences. However, decoding strategies may contribute to the generation of hallucinated text \cite{meister2020if}. Selecting an effective decoding technique can help the model generate output that is more grounded in context and aligned with user expectations. Recent studies have shown promising results in mitigating hallucinations by building upon the on-the-shelf decoding strategies. Chen et al. \cite{chuang2023dola} introduced DoLa to mitigate hallucination in LLMs without the need for fine-tuning or external retrieval mechanisms. DoLa enhances factual accuracy by exploiting differences between mature (higher) and premature (lower) transformer layers. Evaluated on the TruthfulQA benchmark, DoLa attained a 12–17\% improvement in the truthfulness scores in various LLMs. Similarly, Shi et al. \cite{shi2024trusting} enhanced the contextual faithfulness of text generation by reducing the influence of prior knowledge during the decoding process. They proposed a contrastive decoding technique that modifies output probabilities to enhance the influence of the given context. The context-aware decoding strategy operates by contrasting the output probabilities of the model when the context is included and excluded from the prompt. This ensures that the model gives more weight to contextually relevant tokens, which reduces reliance on outdated or incorrect prior knowledge. This approach has been evaluated on a summarization task and reported an enhancement of factuality by 14.3\%. Waldendorf et al. \cite{waldendorf2024contrastive} applied contrastive decoding in multilingual machine translation settings. Their approach maximizes the log-likelihood difference between an expert model and a deliberately source-detached amateur model, which significantly enhances translation fidelity and mitigates hallucinations. Furthermore, Sennrich et al. \cite{sennrich2024mitigating} proposed source-contrastive and language-contrastive decoding methods, contrasting the correct input segment with randomly selected or incorrect segments. This approach substantially reduced severe hallucinations, which is defined by a chrF2 score below 10, by 67–83\% and oscillatory hallucinations by 75–92\% across various language pairs. \textcolor{black}{In order to reduce the response latency of existing methods, Chang et al. \cite{chang2025monitoring} introduced monitoring decoding, which monitors token-by-token generation using a factuality monitor that evaluates partial responses during decoding. When a token or span is predicted to cause hallucination, the model intervenes immediately using a tree-based resampling strategy to revise only the risky tokens instead of regenerating the whole answer. This reduces the number of overconfident hallucinated tokens at their source during generation.}

\textbf{Knowledge Distillation.}
It is a method by which a smaller model, the student, learns to replicate the performance of a larger, well-performing model, the teacher, without significant performance loss \cite{hinton2015distilling}. Recent studies have shown that knowledge distillation is effective in mitigating LLM hallucinations. McDonald et al. \cite{mcdonald2024reducing} used knowledge distillation with the Mistral LLM, demonstrating substantial improvements in factual accuracy and significant reductions in hallucination rates on the MMLU benchmark. The proposed approach used temperature scaling and intermediate layer matching, which enables a compact student model to emulate a larger teacher model. Therefore, the approach ensures more contextually accurate outputs without compromising computational efficiency. Similarly, Nguyen et al. \cite{nguyen2025smoothing} proposed a smoothed knowledge distillation method, where soft labels from a teacher model replaced traditional hard labels, to reduce the model's overconfidence and encourage better factual grounding. This method was evaluated on summarization benchmarks, such as CNN/Daily Mail and XSUM, and it successfully lowered hallucination rates while maintaining robust performance across general natural language processing tasks. Liu et al. \cite{liu2024mind} designed a multi-task learning (MTL) paradigm to distill self-evaluation capabilities from the GPT-3.5-turbo into smaller models such as the T5-base. This approach uses few-shot CoT prompts and self-assessment output to create rationales and pseudo-labels for training. The results in the SVAMP and ANLI datasets demonstrated significant improvements in accuracy. Elaraby et al. \cite{elaraby2023halo} introduced HALO, a framework addressing hallucinations in smaller LLMs such as BLOOM 7B through HALOCHECK, which is a BlackBox knowledge-free metric to evaluate hallucination severity. HALOCHECK uses entailment-based methods to assess consistency, which outperformed existing metrics in detecting contradictions. To mitigate hallucinations, the authors proposed knowledge injection, fine-tuning the model with domain-specific knowledge, and a teacher-student approach in which GPT-4 provided detailed guidance selectively triggered by HALOCHECK. These techniques significantly improved factual consistency and reduced hallucinations, demonstrating HALO's effectiveness in enhancing weak LLMs' reliability in domain-specific tasks.

\textbf{Supervised Fine-Tuning. }
Although supervised fine-tuning can produce hallucinated text, it can be an effective hallucination mitigation strategy when specific factors are carefully considered. Incorporating grounded input into the fine-tuning dataset enables the model to prioritize accurate factual content over speculative knowledge \cite{hu2024mitigating}. For example, if the model frequently hallucinates in text summarization tasks, fine-tuning with a dataset containing grounded content is beneficial, since it guides the model in aligning generated summaries closely with the original content. 

\textit{Uncertainty-aware Learning.} Training the model to associate uncertainty with lower confidence scores is helpful in uncertain situations \cite{li2024know}. For instance, while summarizing a detailed medical article with specialized terms and domain-specific information, the model can be trained to flag areas of uncertainty. As a result, the model may either omit this information from the summary or openly indicate limitations (e.g., "certain details were not specified"). Similarly, instructional supervision can support this approach by teaching the model that disclaiming uncertain information is preferable to generating incorrect data. This method allows the model to respond with "I don't know" when information is unavailable \cite{zhang2023r,zhu2025grait}. \textcolor{black}{Dey et al. \cite{dey2025uncertainty} introduced uncertainty-aware fusion (UAF), which reduces hallucination using an ensemble of multiple LLMs combined through uncertainty estimation. Each model provides both an answer and a confidence signal about the factual correctness. A fusion module then selects or combines responses from models that are both accurate and self-aware. The proposed approach enhanced factuality while also maintaining efficiency.}

\textit{Multi-task Learning.} The hallucination issue may stem from the dependence of the training process on a single dataset, which limits the model's ability to grasp the true features of the task. Incorporating appropriate auxiliary tasks alongside the primary task during fine-tuning helps reduce the model's susceptibility to hallucination issues \cite{ji2023survey}. The process of fine-tuning a model on several tasks concurrently is known as MTL. Learning multiple tasks simultaneously enables models to gather general information beyond task-specific properties. Consequently, learning performance can improve by sharing information across tasks rather than learning each task separately. For example, in an MTL framework, encompassing abstractive summarization and fact-checking makes the model learn to produce coherent summaries and to validate the factual accuracy of the information it generates. The fact-checking task helps to prevent the model from producing erroneous or fabricated statements by explicitly instructing it to assess and verify factual assertions. However, it is essential to carefully select tasks for concurrent learning to avoid the risk of negative transfer, which occurs when learning conflicting features \cite{crawshaw2020multi}.

\textit{Prompt-guided Learning.} Prompt retrieval and selection techniques integrated with fine-tuning have further expanded the capabilities of LLMs to mitigate hallucinations \cite{cheng2023uprise}. Cheng et al. \cite{cheng2023uprise} proposed UPRISE, which is a method of detecting and mitigating hallucinations by enhancing the retrieval of relevant prompts that help the LLM in making more accurate conclusions. The authors fine-tuned a lightweight model to autonomously extract prompts from a pool of prompts based on a zero-shot task input. The model has two encoders: one for processing the task input and the other for processing the prompt. During training, the model maximizes the similarity between task inputs and positive prompts while minimizing the similarity to negative prompts through contrastive learning. UPRISE evaluates retrieval scores according to the LLM's accuracy in anticipating the proper label. Upon fine-tuning the retriever, the fine-tuned retriever is employed to extract the most relevant prompts from the prompt pool for a certain input. The prompts obtained are combined with the task inputs and forwarded to the LLM to produce the final output. The proposed method was evaluated using ChatGPT, and UPRISE outperformed vanilla zero-shot prompting on fact-checking tasks. \textcolor{black}{}


\textbf{Self-learning via Contrastive Learning. }
 Contrastive learning is a self-supervised technique that aims to acquire valuable data representations by differentiating between positive and negative samples. The fundamental concept is to reduce the proximity of representations of similar positive pairs while increasing the distance between representations of dissimilar pairs \cite{chopra2005learning}. In hallucination mitigation, contrastive learning enhances models' ability to acquire precise representations by focusing on what makes data points factually or contextually consistent. By training on positive pairs (e.g., grounded factual content) and negative pairs (e.g., mismatched or hallucinated content), a model can better distinguish between factual and non-factual information \cite{ji2023survey}. For example, contrastive learning can be applied to train a model for text summarization, with positive samples representing reference summaries and negative samples indicating hallucinated summaries. Contrastive learning helps to differentiate between them, thereby aiding in the reduction of hallucination \cite{cao2021cliff}. Sun et al. \cite{sun2023contrastive} introduced MixCL, a mixed contrastive learning approach specifically designed to mitigate hallucination in conversational language models. By employing a combination of hard negative sampling strategies and span-level mixing of positive and negative knowledge samples, MixCL significantly improved the models' abilities to distinguish factual content from hallucinated information. However, the selection of positive and negative pairs, while beneficial, must be monitored to prevent the emergence of suboptimal representations.

\textbf{Cross-Lingual Learning. }
Cross-lingual transfer learning enables LLMs to exploit knowledge that is richer or more complete in one language, often English, to generate responses in another low-resource language. Pre-training a bi-lingual or multi-lingual model, then fine-tuning it on limited data from a low-resource language can speed up model convergence and improve performance \cite{abdelrahman2024hallucination}. However, it can also increase the risk of hallucination if the data contains a language mismatch or noisy tokens. Cross-lingual learning can contribute to mitigating hallucination in the pre-training, fine-tuning, and inference stages.

During cross-lingual continual pre-training, noisy tokens, such as artifacts and emojis, can distort the learned distribution of the data in the new language, which leads to hallucination. Fan et al. \cite{fan2025weighted} proposed InfoLoss, which augments cross-entropy with normalized pointwise mutual information weights over local context. This is to down-weight tokens with low co-occurrence support and thereby limit their influence. Continual pre-training Llama-2-7B with InfoLoss over a bi-lingual corpus improved cross-lingual transfer on 12 benchmarks and reduced hallucinations relative to size-matched baselines. Zheng et al. \cite{zheng2025ccl} developed curriculum-based contrastive learning, which aligns multilingual semantic spaces during continued pre-training.

Qiu et al. \cite{qiu2023detecting} proposed mFACT, a multilingual factuality metric designed to evaluate hallucinations beyond English. Instead of building separate evaluators for each language, mFACT leverages existing English factuality metrics by translating their supervision signals into the target language. When applied to summarization tasks, incorporating mFACT into loss-weighted training allowed the model to down-weight unfaithful examples, which reduced hallucinations and improved overall summary quality across six languages. Zheng et al. \cite{zheng2025ccl} developed a cross-lingual COT that reasons in a high-resource language before producing the final answer in the target low-resource language.

Huang et al. \cite{huang20241} developed a low-resource knowledge detector that flags the query contents that are written in a low-resourced language. The system then selects a suitable high-resource language, translates the query into the selected language, generates a response in the selected language, and replaces or integrates the answer back into the original language. Experiments on six LLMs and five bilingual datasets show performance gains and reduced cross-language disparities.

\subsection{Mitigation Challenges}
Hallucination mitigation techniques aim to prevent LLMs from producing responses that are factually inaccurate, ungrounded, or contextually inconsistent. 
Most hallucination mitigation approaches proposed in the literature fall into two categories: retrieval and reasoning-based. Retrieval-based approaches leverage external knowledge, while reasoning-based approaches employ logical, step-by-step reasoning to improve the factual grounding, contextual relevance, and overall accuracy of LLM outputs.
Despite substantial progress in retrieval-based hallucination mitigation, these approaches still face several challenges. The effectiveness of these methods is fundamentally constrained by the coverage, timeliness, and quality of the external knowledge sources. Therefore, if relevant information is missing, outdated, or inconsistent, hallucination mitigation may fail or even introduce new errors \cite{zhang2025hallucination}. Furthermore, integrating retrieval results with LLMs often introduces additional latency and complexity into the generation pipeline, which can limit scalability and real-time applicability \cite{agrawal2024can}.

Similarly, reasoning-based mitigation techniques, such as CoT, self-consistency, and iterative refinement, have some limitations. 
The computational cost of deploying reasoning-based methods can be substantial, since these methods depend on generating multiple reasoning chains or performing step-by-step refinements to mitigate hallucination \cite{cheng2025think}. Moreover, iterative and self-correction strategies cannot fully guarantee error removal, even when supported by self-verification and multi-turn feedback \cite{lin2025zebralogic}. Additionally, if initial reasoning chains are flawed, subsequent refinements may amplify rather than correct errors, which underscores the fragility of self-improvement strategies when the underlying logic is unsound \cite{varshney2024investigating}.



Mitigating hallucination through designing well-crafted prompts has been investigated in some studies. Although prompt designs effectively guide the model to the desired output, they have several drawbacks. Designing and refining high-quality domain-specific prompts that maximize accuracy and minimize errors can be resource-intensive, requiring both human expertise and iterative testing. Prompt designs may also be constrained to specific tasks, which limits their generalization to a wide range of applications. While recent advances in automatic prompt optimization, such as prompt tuning \cite{lester2021power, liu2022p} and prefix-tuning \cite{li2021prefix}, show promise in reducing manual effort, integrating these approaches with hallucination mitigation techniques needs to be investigated.


Other researchers have proposed some enhancements to the model architecture or fine-tuning strategies to mitigate hallucinations in the LLMs outputs. 
Despite notable progress in model development and fine-tuning strategies for hallucination mitigation in LLMs, several challenges remain. Decoding strategies, such as contrastive, source-aware, and DoLa methods, have improved contextual faithfulness but can still struggle with generalization across domains and sensitivity to parameter tuning \cite{chuang2023dola, shi2024trusting, sennrich2024mitigating}. Knowledge distillation approaches are constrained by the quality of teacher models and can inadvertently transfer biases or hallucinations. Moreover, their reliance on computationally intensive methods and high-quality data can limit scalability, especially for multilingual or specialized domains \cite{mcdonald2024reducing, nguyen2025smoothing}. Supervised fine-tuning requires large, reliably annotated datasets, which are often unavailable or costly to produce for low-resource tasks. Furthermore, MTL can result in negative transfer and overfitting if auxiliary and primary tasks are not well-aligned \cite{hu2024mitigating, ji2023survey}. Contrastive learning is powerful for distinguishing factual from hallucinated content, but is sensitive to the choice and quality of positive and negative samples \cite{sun2023contrastive, cao2021cliff}. Cross-lingual learning offers promise but suffers from increased hallucination rates in low-resource or syntactically divergent languages, which requires extensive resources and careful parameter sharing \cite{hu2020xtreme, ansell2022composable}.

\textcolor{black}{The observed limitations across mitigation strategies indicate that no single approach can address all hallucination failure modes, underscoring the need for domain-aware, task-adaptive pipelines that combine complementary techniques. Retrieval-based approaches tend to be more beneficial in knowledge-intensive domains, such as medicine, law, and finance. However, retrieval-based mitigation is less reliable in low-resource languages or rapidly evolving domains, where trustworthy sources are scarce or inconsistent. In contrast, reasoning-based strategies are more helpful for multi-step reasoning tasks, such as math logical QA, but may not be useful in domains where hallucinations arise from missing or incorrect world knowledge rather than from flawed reasoning chains. Prompt-based and decoding-based mitigation techniques often show gains in structured or well-defined tasks, such as summarization or information extraction, where instructions can more tightly control faithfulness to input. However, their effectiveness declines in open-ended generation settings, where hallucinations arise from broad knowledge gaps rather than instruction-following failures. Similarly, model-centric approaches such as fine-tuning, contrastive learning, and distillation tend to perform well in domain-specific deployments with curated training data, but they struggle to generalize to unseen domains and may even amplify domain-specific biases or misconceptions. This highlights the need for domain-aware, task-adaptive mitigation pipelines that combine retrieval, reasoning control, and model-level adaptation tailored to the deployment setting.}

\begin{table*}[t]
\centering
\footnotesize
\setlength{\tabcolsep}{5pt}
\scriptsize
\caption{Summary of the surveyed hallucination mitigation techniques. 
Columns indicate whether the approach uses prompt engineering (P), retrieval (R), self-refine/reasoning (S), or model-centric (M).}
\label{tab:mitigation-comparison}

\resizebox{\textwidth}{!}{

\begin{tabular}{@{} l c >{\centering\arraybackslash}p{2.5cm} 
p{0.3cm} p{0.3cm} p{0.3cm} p{0.3cm} 
p{3.5cm} p{4.5cm} @{}}

\toprule

\multirow{2}{*}{\textbf{Ref}} 
& \multirow{2}{*}{\textbf{Year}} 
& \multirow{2}{*}{\textbf{Task / Domain}} 
& \multicolumn{4}{c}{\textbf{Mitigation Method}} 
& \multirow{2}{*}{\textbf{Dataset(s)}} 
& \multirow{2}{*}{\textbf{Metric(s)}} \\

\cmidrule(lr){4-7}

& & 
& \textbf{P} 
& \textbf{R} 
& \textbf{S} 
& \textbf{M} 
& & \\

\midrule

\cite{jiang2023ai} & 2023 & News generation & \cmark & & & \cmark & Private & MAUVE, Topic consistency, Core consistency \\

\cite{yang2023refgpt} & 2023 & Dialogue & \cmark & \cmark & & & RefGPT-Fact, RefGPT-Code & Human evaluation, LLM-as-a-judge \\

\cite{kim2024self} & 2024 & Legal instruction gen & \cmark & \cmark & & \cmark & Private & Accuracy, Fluency \\

\cite{feldman2023trapping} & 2023 & QA citation & \cmark & & & & Private & Accuracy \\

\cite{zhang2023summit} & 2023 & Summarization & \cmark & & \cmark & \cmark & CNN/DM, XSum, NEWTS & Rouge, G-eval, Human \\

\cite{vu2023freshllms} & 2023 & QA & \cmark & \cmark & & \cmark & FRESHQA & Accuracy \\

\cite{penkov2024mitigating} & 2024 & QA, Summarization & \cmark & \cmark & & & Private & -- \\

\midrule

\cite{nathani2023maf} & 2023 & QA & & \cmark & \cmark & \cmark & DROP, GSM-IC, EntailmentBank & Accuracy \\

\cite{huo2023retrieving} & 2023 & QA & \cmark & \cmark & \cmark & & MS MARCO & Accuracy, Human \\

\cite{li2024traq} & 2024 & QA & & \cmark & & \cmark & NQ, TriviaQA, SQuAD, BioASQ & Coverage rate \\

\cite{ding2024retrieve} & 2024 & QA & & \cmark & \cmark & \cmark & TruthfulQA, StrategyQA, NQ, TriviaQA & GPT-judge, BLEU, Rouge \\

\cite{tian2024kg} & 2024 & QA & & \cmark & & \cmark & OBQA, CSQA, WQSP, CWQ & Accuracy, Hits@1 \\

\cite{ji2024retrieval} & 2024 & QA & & \cmark & \cmark & \cmark & WebQSP, CWQ & F1, Hits@1 \\

\cite{baek2023knowledge} & 2023 & QA & \cmark & \cmark & & & WebQuestionsSP, Mintaka & MRR, Top-K Accuracy \\

\cite{trivedi2023interleaving} & 2023 & QA & & \cmark & \cmark & & HotpotQA, 2WikiMultihopQA & Recall, F1 \\

\cite{guan2024mitigating} & 2024 & QA & & \cmark & \cmark & \cmark & Mintaka, HotpotQA & EM, F1 \\

\cite{yang2025curiousllm} & 2025 & QA & & \cmark & \cmark & & Follow-up QA & Accuracy, EM \\

\cite{tan2025paths} & 2025 & QA & & \cmark & \cmark & \cmark & CWQ, WebQSP, GrailQA & EM \\

\midrule

\cite{gao2023rarr} & 2023 & QA, Dialogue & \cmark & \cmark & \cmark & \cmark & NQ, SQA, QReCC & Human, Attrauto \\

\cite{song2024rag} & 2024 & QA, Summarization & \cmark & \cmark & & \cmark & RAGTruth & Precision, Recall, F1 \\

\cite{sultan2024structured} & 2024 & Dialogue & \cmark & \cmark & \cmark & \cmark & DoQA, QuAC & F1, Accuracy \\

\cite{huang2024coq} & 2024 & QA & & \cmark & \cmark & & HotpotQA & F1, Human \\

\cite{zhang2024chain} & 2024 & QA & & \cmark & \cmark & \cmark & CommonsenseQA, ARC & EM \\

\cite{li2023chain} & 2023 & QA & \cmark & \cmark & \cmark & \cmark & FEVER, HotpotQA & Accuracy, EM \\

\cite{yu2023chain} & 2024 & QA & & \cmark & \cmark & \cmark & NQ, TriviaQA & F1, EM \\

\cite{cheng2025think} & 2025 & QA & & & \cmark & \cmark & TruthfulQA, HaluEval & Accuracy \\

\midrule

\cite{waldendorf2024contrastive} & 2023 & Machine translation & & & & \cmark & FLORES-101 & COMET \\

\cite{sennrich2024mitigating} & 2024 & Machine translation & & & & \cmark & HLMT & SpBLEU, ChrF2 \\

\cite{shi2024trusting} & 2024 & Summarization & & & & \cmark & CNN-DM, XSUM & ROUGE-L, FactKB \\

\cite{nguyen2025smoothing} & 2025 & Summarization & & & & \cmark & CNNDM, XSUM & ROUGE-L \\

\cite{cao2021cliff} & 2024 & Summarization & \cmark & & & \cmark & XSum, CNN/DM & QuestEval, FactCC \\

\cite{qiu2023detecting} & 2023 & Summarization & & & & \cmark & XLSum & mFact, Rouge \\

\cite{chuang2023dola} & 2023 & QA & \cmark & & \cmark & \cmark & TruthfulQA & Accuracy \\

\cite{elaraby2023halo} & 2023 & QA & \cmark & \cmark & \cmark & \cmark & Private & HALOCHECK \\

\cite{hu2024mitigating} & 2024 & QA & \cmark & \cmark & & \cmark & HaluEval, TruthfulQA & Accuracy \\

\cite{chang2025monitoring} & 2025 & QA & & & & \cmark & TruthfulQA, TriviaQA & Accuracy \\

\cite{dey2025uncertainty} & 2025 & QA & & & & \cmark & TruthfulQA, TriviaQA & Accuracy \\

\cite{sun2023contrastive} & 2023 & Dialogue & & \cmark & & \cmark & WoW & F1, Rouge, BLEU \\

\cite{cheng2023uprise} & 2024 & Multiple tasks & \cmark & \cmark & & \cmark & Multiple datasets & Accuracy \\

\bottomrule

\end{tabular}
}

\end{table*}
\section{Benchmark Datasets}
\label{sec_benchmark}
Although LLM hallucination detection and mitigation is a recent research area, substantial work has been directed toward curating datasets for detecting and mitigating hallucination in varied NLG tasks. Hallucination datasets can be used for hallucination detection, mitigation, or both. Hallucination detection datasets are typically paired inputs with outputs, with explicit annotation of instances of hallucination that allow researchers to assess a model's tendency toward producing false or fabricated information. On the other hand, hallucination mitigation datasets place an emphasis on providing high-quality, factually correct, and contextually grounded input-output pairs. They often include external references or grounding information, such as retrieval-based support documents that foster models to produce outputs that are more reliable and faithful. 

Hallucination detection datasets can be used for mitigation by repurposing the annotated data to guide model training or refinement processes that directly address hallucination. They can be used to fine-tune the model with explicit negative examples, create contrastive learning tasks, or analyze hallucinated outputs to design more effective prompts. It can also be integrated into the generation pipeline to flag or reject hallucinated outputs before presenting them to the user. Table \ref{Tab:mit} compares the publicly available datasets for hallucination detection and mitigation.  

Table 4 indicates that QA and summarization NLG tasks are predominant, being the principal tasks for which hallucination detection and mitigation have been evaluated. English remains the leading language in detecting and mitigating hallucination. This is due to its high resources in several domains. Other languages are starting to emerge, such as German (WMT18, Absinth), Arabic (Halwasa), Chinese (UbgEval), and multi-lingual datasets (HalOmi). Moreover, it is shown that most of the proposed datasets support both hallucination detection and mitigation experiments, which provide reliable benchmarks for evaluating different techniques. In addition, the ground-truth labels are predominantly derived from human evaluation or human-curated references, ensuring high-quality factual annotations even when the initial source material comes from Wikipedia, news articles, or model-generated text.

\textcolor{black}{Tables \ref{tab:detection-comparison} and \ref{tab:qa_sota} show that hallucination detection and mitigation remain highly task- and benchmark-dependent. No single detection or mitigation method consistently outperforms others across all datasets. The performance of detection and mitigation techniques fluctuates under domain and task shifts. Furthermore, the diversity of benchmarks and evaluation setups complicates direct comparisons, which underscores the need for standardized evaluation protocols and hybrid approaches tailored to specific task characteristics.} 

\textcolor{black}{While most existing benchmarks focus on evaluating the performance of hallucination detection and mitigation methods rather than measuring how frequently hallucinations occur in LLMs, recent community efforts provide useful empirical data on hallucination rates across systems. One such resource is the Hallucination Leaderboard \cite{vectara_hallucination_leaderboard}, which aggregates model performance on grounded summarization tasks and reports hallucination rates computed by the Hughes Hallucination Evaluation Model (HHEM) \cite{vectara_hallucination_leaderboard}. Some state-of-the-art models achieve hallucination rates below a few percentage points on this task. This indicates that in narrow, constrained settings, such as summarization, hallucinations are relatively infrequent compared with unconstrained generation. However, substantial variation remains across model families and prompt behaviors, which demonstrates that hallucination is neither uniform nor negligible even under strict grounding conditions. While such task-specific benchmarks do not yet provide a unified, cross-task hallucination frequency metric, the emerging leaderboard results represent one of the first systematic efforts to quantify hallucination prevalence and offer the community a common reference point for comparing models on factual consistency.}

\begin{table*}[t!]
\centering
\tiny
\caption{Benchmark Datasets for Hallucination Detection and Mitigation.}%
\label{Tab:mit}

\begin{tabularx}{\textwidth}{
   p{0.03\textwidth}
   p{0.09\textwidth}
   p{0.03\textwidth}
   p{0.06\textwidth}
   p{0.05\textwidth}
   p{0.06\textwidth}
   p{0.07\textwidth}
   p{0.12\textwidth}
   p{0.11\textwidth}
   p{0.07\textwidth}
}
\toprule
\textbf{Ref} & \textbf{Name} & \textbf{Year} & \textbf{Lang.} &
\textbf{Detect} & \textbf{Mitigate} &
\textbf{Domain} & \textbf{Source} & \textbf{Size} & \textbf{GT Source} \\
\midrule

\cite{vu2023freshllms} & FreshQA & 2023 & \multirow{12}{*}{English} & \cmark & \cmark & QA & Human & 600 & Human  \\
\cite{kasai2023realtime} & RealtimeQA & 2022 &  & \cmark & \cmark & QA, MCQ & Human & Dynamic & Human  \\
\cite{goodrich2019assessing} & WikiFact & 2019 &  & \cmark &  & Summ. & Wikipedia & 36.9M & Wikipedia \\
\cite{xsum-emnlp} & XSum & 2020 &  & \cmark & \cmark & Summ. & BBC & 226,711 & BBC article \\
\cite{yang2018hotpotqa} & HotpotQA & 2018 &  & \cmark & \cmark & QA & Wikipedia & 112,779 & Human  \\
\cite{li2023halueval} & HaluEval & 2023 &  & \cmark & \cmark & QA, Summ., Dialogue & Human, Model-gen & 35,000 & Human  \\
\cite{rahman2024defan} & DefAn & 2024 &  & \cmark & \cmark & QA & Human & 68K public, 7.5K hidden & Human  \\
\cite{parikh2020totto} & ToTTo & 2020 &  & \cmark & \cmark & Data2Text & Wikipedia tables & 136,161 & Human  \\
\cite{joshi2017triviaqa} & TriviaQA & 2017 &  & \cmark & \cmark & QA & Wikipedia and web search & 78.8K (Wiki); 95K (Web); 1,975 (Clean) & Human  \\
\cite{gupta2022dialfact} & DialFact & 2022 &  & \cmark & \cmark & Dialogue & WoW convos, Wikipedia & 22,245 & Human, Wikipedia \\
\cite{kryscinski2020evaluating} & FactCC & 2020 &  & \cmark &  & Summ. & CNN/DailyMail & $>$1M & Human  \\

\cite{pandit2025medhallu} & MedHallu & 2025 &  & \cmark & \cmark & QA & PubMedQA & 10K & PubMedQA \\
\midrule

\cite{bojar-EtAl:2018:WMT1} & WMT18 & 2018 & German-English & \cmark & \cmark & MT & News translations & 3.4K annotations, 1.3M sentences & Human  \\
\cite{mascarell2024german} & Absinth & 2024 & German & \cmark &  & Summ. & 20Minuten news, model-gen & 4,314 & Human  \\
\midrule

\cite{mubarak2024halwasa} & Halwasa & 2024 & \multirow{4}{*}{Arabic} & \cmark & \cmark & Text gen. & Model-gen & 10K & Human  \\

\cite{mubarak2025islamiceval} & IslamicEval & 2025 &  & \cmark & \cmark & QA &  & 1,506 &  \\
  
\cite{alansari2025arahallueval} & AraHalluEval & 2025 &  & \cmark &  & QA, Summ. &  & 200QA, 100Sum. & Human \\
  
\cite{mohammed2025aftina} & Aftina & 2025 &  &  & \cmark & QA &  & 18k &  \\
\midrule

\cite{liang2024uhgeval} & UHGEval & 2024 & \multirow{2}{*}{Chinese} & \cmark & \cmark & Text gen. & Chinese news & 5,141 & GPT-4, Human  \\
\cite{cheng2023evaluating} & HaluQA & 2023 &  & \cmark & \cmark & QA & Human, Model-gen & 450 & Human  \\
\midrule

\cite{pal2023med} & Med-Halt & 2023 & \multirow{3}{*}{Multi} & \cmark & \cmark & MCQ QA & Human, Model-gen & 4,916 & Human  \\
\cite{dale2023halomi} & HalOmi & 2023 &  & \cmark &  & MT & NLLB-200 translations, Wikipedia & $\sim$3.5K--4K & Human  \\

\cite{vazquez2025semeval} & Mu-Shroom & 2025 &  & \cmark &  & General & Wikipedia, Model-gen & $\sim$5.7K & Human  \\
\bottomrule
\end{tabularx}
\end{table*}

\begin{table*}[ht]
\centering
\scriptsize
\caption{\textcolor{black}{Benchmark results for hallucination mitigation techniques using the most utilized datasets.}}
\label{tab:qa_sota}
\begin{tabular}{l |cc|cc|ccc|c|cc|c}
\toprule
\textbf{Ref} 
& \multicolumn{2}{c|}{\textbf{TriviaQA}} 
& \multicolumn{2}{c|}{\textbf{NQ}} 
& \multicolumn{3}{c|}{\textbf{HotpotQA}} 
& \multicolumn{1}{c|}{\textbf{XSum}} 
& \multicolumn{2}{c|}{\textbf{TruthfulQA}} 
& \multicolumn{1}{c}{\textbf{CNN/DM}} \\

& EM & F1 
& EM & F1 
& RL & EM & F1 
& RL 
& RL & Acc 
& RL \\

\hline
\cite{zhang2023summit} &  &  &  &  &  &  &  & 17.6 &  &  & 26.9 \\
\cite{ding2024retrieve} & 69.0 & 78.5 & 40.0 & 57.3 &  &  &  &  & 31.2 &  &  \\
\cite{trivedi2023interleaving} &  &  &  &  & 49.3 & 60.7 &  &  &  &  &  \\
\cite{guan2024mitigating} &  &  &  &  & 34.0 & 47.2 &  &  &  &  &  \\
\cite{gao2023rarr} &  &  &  & 57.0 &  &  &  &  &  &  &  \\
\cite{huang2024coq} &  &  &  &  &  &  & 90.3 &  &  &  &  \\
\cite{li2023chain} &  &  &  &  &  & 35.4 &  &  &  &  &  \\
\cite{shi2024trusting} &  &  &  &  &  &  &  & 20.0 &  &  & 27.1 \\
\cite{nguyen2025smoothing} &  &  &  &  & 7.4 &  &  & 20.0 &  &  & 27.8 \\
\cite{cao2021cliff} &  &  &  &  &  &  &  & 38.2 &  &  &  \\
\cite{chang2025monitoring} & 80.8 &  & 47.4 &  &  &  &  &  &  &  &  \\
\cite{cheng2025think}&  &  &  &  &  &  &  &  &  & 45.1 &  \\
\cite{dey2025uncertainty}&  &  &  &  &  &  &  &  &  & 48.4 &  \\
\cite{chang2025monitoring}&  &  & 31.0 &  &  &  &  &  &  & 50.2 &  \\
\cite{yu2023chain}& 76.3 & 82.6 & 57.5 & 48.9 &  &  &  &  &  & 50.2 &  \\
\bottomrule
\end{tabular}
\end{table*}

\section{Hallucination Detection and Mitigation Metrics}
\label{sec_metric}
Hallucination evaluation metrics are essential to objectively measure the truthfulness of LLMs on different NLG tasks. Hallucination evaluation metrics can be categorized into statistical, data-driven, human-based, and mixed metrics \cite{narayanan-venkit-etal-2024-audit}. Each provides unique insights and exhibits limitations in terms of faithfulness, factuality, and generalizability. 

\textit{Statistical metrics} compute hallucination scores by measuring mismatches between the generated content and references \cite{ji2023survey}. Statistical metrics include token overlap metrics, such as BLEU and Rouge, semantic similarity metrics, such as BERTScore, and uncertainty-based metrics, such as perplexity. These approaches are widely adopted due to simplicity, but often lack robustness in aligning with human judgments of hallucination \cite{maynez2020faithfulness}. 

\textit{Data-driven metrics} use curated datasets or other models' outputs for hallucination detection by measuring content mismatching. SelfCheckGPT \cite{manakul2023selfcheckgpt} is a reference-free metric that evaluates consistency across multiple LLM outputs. By generating and comparing paraphrased outputs, it identifies unsupported information. Similarly, FactCC \cite{kryscinski2020evaluating} is a supervised NLI classifier trained on synthetically perturbed data to detect factual inconsistencies between source and generated text. G-eval \cite{liu2023gevalnlgevaluationusing} is an LLM-as-a-judge evaluation framework that uses CoT prompting. It transforms user-defined evaluation criteria into structured reasoning steps, which are then executed by LLMs, typically GPT-4, to score outputs in a form-filling paradigm. Evaluation using LLMs like GPT-4 achieves the best overall alignment with human judgment \cite{kulkarni2025evaluatingevaluationmetrics}. Moreover, mFACT \cite{qiu2023detecting} enhances cross-lingual hallucination detection by translating non-English outputs into English and applying multiple English-trained faithfulness metrics.



\textit{Human annotation} remains a gold standard method for assessing hallucination. Several studies employ human annotators to detect LLM hallucination \cite{maynez2020faithfulness,huo2023retrieving,huang2024coq,elaraby2023halo,cao2021cliff,alansari2025arahallueval}. Human evaluation is typically conducted along multiple dimensions such as faithfulness, the degree to which generated content aligns with source facts; factuality, the correctness of the information with respect to real-world knowledge or verified ground truth; consistency, absence of contradictions within the output; relevancy,  appropriateness of the response to the input query; and adequacy, completeness of the conveyed information. Another human-based technique to detect hallucination is by using eye tracking, where a reader's gaze patterns, such as prolonged fixations, regressions, and pupil dilation, are monitored to identify text segments that cause unexpected cognitive load and may therefore contain hallucinated or unfaithful content. \cite{maharaj-etal-2023-eyes}. However, this approach is time-consuming and resource-intensive \cite{maynez2020faithfulness}.

To harness the strengths of automated and manual assessments, \textit{mixed approaches,} including factor analysis of mixed data (FAMD) \cite{kulkarni2025evaluatingevaluationmetrics}, and advanced datasets including HalluLens \cite{bang-etal-2025-hallulens}, combine multiple signals, such as semantic similarity, QA-based judgment, and consistency scores, with human annotations or domain expertise. These methods provide comprehensive, context-aware analysis of hallucinations across varied domains and use cases.

\section{Open Issues and Future Directions}
\label{sec_open_issues}
Despite a large number of previously proposed approaches, a multitude of challenges remain unsolved in the LLMs'
hallucination detection and mitigation methods. This section categorizes these issues into detection, mitigation, and resource issues. 

\subsection{Hallucination Detection}

\vspace{2pt}\noindent\textbf{Generalization.} Most of the current hallucination detection methods are trained and evaluated on particular datasets and tasks, resulting in inadequate generalization across diverse datasets and tasks. Additionally, the inconsistency in model training and prompting techniques hinders the development of universal hallucination detection frameworks \cite{binkowski2025hallucination}.  As a result, detecting the hallucination of different LLMs will require adapting multiple methods based on the characteristics of the target LLM and downstream task, which complicates real-time and resource-limited deployment.  

\vspace{2pt}\noindent\textbf{Computational Overhead.} Hallucination detection techniques, such as self-consistency checks, uncertainty estimation, and retrieval-augmented verification, often require multiple inference iterations or access to external databases, thereby escalating computational expenses. This poses challenges for real-time applications and deployment in resource-limited settings. Although AGSER \cite{liu2025attention} reduces computational costs compared to current methods, it still requires several inference iterations. Future research should focus on developing efficient and lightweight detection mechanisms. Techniques such as knowledge distillation, quantization, or sparse architectures could help reduce computational overhead without compromising detection accuracy.


\vspace{2pt}\noindent\textbf{Subtle Hallucination Detection. } Subtle factual errors in LLM-generated responses are often difficult to detect and may require domain expertise or large external resources for identification. Future research should explore methods such as diffusion-based contrastive learning models to generate both factual and hallucinated versions of an answer. By learning the semantic distance between them, a model could develop a more nuanced understanding of hallucinations.

\vspace{2pt}\noindent\textbf{Multi-Turn Dialogue and Long-Form Generation.} Detecting hallucinations in conversational agents is more challenging than in single-turn dialogues due to context propagation across multiple turns. Hallucinations may emerge gradually, making post-hoc detection less effective. Future research should focus on techniques such as dynamic context tracking, memory-augmented models, and reinforcement learning with hallucination-specific rewards to address this issue.


\vspace{2pt}\noindent\textbf{Low-Resource Languages.} Most hallucination detection methods are developed and evaluated on high-resource languages like English, leaving low-resource languages underrepresented. This limits the applicability of hallucination detection methods in multilingual and global contexts. Future work should focus on extending hallucination detection to low-resource languages by leveraging cross-lingual transfer learning, multilingual pretraining, and data augmentation techniques. Additionally, creating multilingual datasets and benchmarks would enable more comprehensive evaluation and development of detection methods for diverse linguistic settings.

\subsection{Hallucination Mitigation}

\vspace{2pt}\noindent\textbf{Limitations of Attention Mechanisms. }
Attention mechanisms, while being the core of transformer architectures, usually fail to properly distinguish between useful context and noisy information due to the softmax bottleneck. Current attention mechanisms may improperly weigh context and self-generated content, causing contextually irrelevant hallucinations \cite{huang2025dynamic}.  Future work should investigate enhancing attention mechanisms to dynamically emphasize key contexts with adaptive uncertainty measures, boosting context faithfulness, and reducing hallucination \cite{oorloff2025mitigating}.

\vspace{2pt}\noindent\textbf{Suboptimal Exploration in Reasoning Tasks. } The current exploration strategies, such as prompt-based Monte Carlo Tree Search, lack adaptive adjustment, which leads to either insufficient or excessive exploration. This imbalance can result in premature convergence or overlooking correct reasoning pathways, thus failing to adequately mitigate hallucinations \cite{duan2025prompt}. Future research should focus on adaptive exploration strategies, possibly integrating dynamic threshold adjustments or feedback-based exploration policies, to balance exploration and exploitation effectively.

\vspace{2pt}\noindent\textbf{Cross-Lingual and Multilingual Challenges. }
Current cross-lingual and multilingual models face some challenges in maintaining factual consistency across languages, particularly in low-resource settings. Multilingual LLMs typically do not cross-learn across languages, especially for implicit reasoning tasks. For instance, models can correctly answer questions in English but fail in Swahili even with equivalent knowledge. Moreover, more training favors high-resource languages, but it leads to unreliable outputs in low-resource languages \cite{chua2025crosslingualcapabilitiesknowledgebarriers}. Future work should focus on developing adaptive multilingual models that can dynamically scale representations based on linguistic context.

\vspace{2pt}\noindent\textbf{Low-Resource Languages.}
Current hallucination mitigation techniques perform poorly in low-resource languages due to insufficient data and limited linguistic coverage. Future research should focus on methods tailored to low-resource languages, such as few-shot cross-lingual transfer learning that leverages high-resource languages to improve performance. In addition, multilingual knowledge distillation should be explored to facilitate knowledge transfer from high-resource to low-resource models.

\vspace{2pt}\noindent\textbf{Fine-tuning limitations.} Fine-tuning LLMs using traditional input-output pairs is more likely to lead to overfitting, catastrophic forgetting, and increased hallucinations, especially when training data is noisy or biased. Fine-tuning without critiques or adaptive weighting can lead to over-refusal or continued hallucinations \cite{zhu2025grait}. Future research should investigate critique-based tuning and uncertainty-weighted adaptive fine-tuning techniques for LLMs hallucination mitigation.

\subsection{Benchmarks and Evaluation Metrics}
\vspace{2pt}\noindent\textbf{Benchmark Coverage and Diversity.} Many existing benchmarks are related to specific tasks, such as QA, summarization, or translation, and are limited to particular sources, including news and Wikipedia. This introduces task and domain bias, limiting generalization to open-ended or multi-domain generation scenarios. Moreover, benchmarks for dialogue and code generation hallucination are still scarce, which limits the applications in these domains. 

\vspace{2pt}\noindent\textbf{Binary or Coarse-Grained Labels.} Many hallucination detection benchmarks reduce the output to a binary label—“hallucinated” vs “non-hallucinated.” This approach leads hallucination detection methods to ignore partial hallucinations and varying error severity. Therefore, subtle factual drifts are not distinguished, which limits diagnostic granularity and reduces the benchmark’s value for model improvement.

\vspace{2pt}\noindent\textbf{Dependence on Human Annotation.} Ground truths in many benchmarks depend heavily on human annotators to label hallucinated spans or judge factual correctness. This introduces several issues related to scalability, inter-annotator agreement, and bias. 

\vspace{2pt}\noindent\textcolor{black}{\textbf{Bias of LLM-as-Judge.}  Recent evidence shows that LLM-as-a-judge can be influenced by visually appealing formatting, which leads to large preference shifts even when the underlying content quality is unchanged \cite{chen2024humans}. This is particularly concerning for open-ended evaluation where judges may be convinced by perceived credibility rather than factual correctness \cite{chen2024humans}. In addition, LLM-as-a-judge exhibits positional effects. The same judge may prefer different candidates when the order of candidates is swapped, and this behavior is not explained by random variation \cite{shi2025judging}. LLM-as-a-judge also suffers from cognitive-related biases. Therefore, it is recommended to avoid using the same model to both generate and judge outputs \cite{li2024llms}. Future research should investigate using jury-style aggregation across multiple diverse judges to reduce self-enhancement and idiosyncratic preferences~\cite{li2024llms,chehbouni2025neither}. It is also recommended to test the judges using format perturbations, fake-reference perturbations, and adversarial phrasing tests as part of the evaluation protocol.} 

\vspace{2pt}\noindent\textbf{Lack of Explainability and Rationale.} Most benchmarks consist of only a predicted label without an accompanying explanation or rationale. As a result, detection and mitigation methods lack transparency, which makes it hard to trace why a decision was made. 

\vspace{2pt}\noindent\textbf{Cross-Lingual and Low-Resource Language Evaluation.} Most existing benchmarks and evaluation metrics focus on high-resource languages, especially English, which leaves many languages underrepresented. Therefore, models may perform well under English benchmarks but fail catastrophically in other languages.

\section{Conclusion}
\label{sec:conclusion}
Hallucination in LLMs remains a critical challenge that undermines the credibility and reliability of AI-generated content across applications. This survey offers a comprehensive review of the underlying causes of hallucination and proposes a taxonomy of five dimensions for detection techniques and four dimensions for mitigation strategies. We further review the benchmarks and performance metrics used in hallucination detection and mitigation techniques, and highlight current limitations and promising directions for future research to foster more factual, trustworthy LLMs.

Our review revealed that despite significant progress made in understanding, detecting, and mitigating hallucination, there remain some significant challenges. Current methods frequently exhibit limitations in generalizability, computational efficiency, handling of multilingual contexts, and interpretability. Directions for future research must focus on the development of robust and more generalizable hallucination detection methods that can be easily applied across multiple domains and model architectures. Improving the interpretability of detection and mitigation techniques will be crucial for building trust in AI systems. Additionally, there is a need for standardized evaluation frameworks and benchmarks that can comprehensively assess hallucination across different tasks and languages. Future work will also need to concentrate on enhancing attention mechanisms, refining exploration strategies in reasoning tasks, raising cross-lingual and low-resource applicability, and enhancing fine-tuning techniques to balance generalization and specificity to better mitigate hallucination. Besides, developing comprehensive benchmarks and datasets to enable detailed evaluation and promoting scalable, interpretable approaches will be essential in the advancement of this field. 

As LLMs continue to evolve, addressing hallucination will remain a critical area of study. By combining advances in model architecture, training techniques, and external knowledge integration, researchers can work towards creating more reliable and factually grounded LLMs. Ultimately, mitigating hallucination is critical to unleashing the full potential of LLMs in applications and making sure that their output remains accurate, reliable, and human values-aligned.

\IEEEdisplaynontitleabstractindextext
\ifCLASSOPTIONcaptionsoff
  \newpage
\fi



%
\bibliographystyle{IEEEtran}
\bibliography{references}

%








\end{document}